\documentclass[10pt,twocolumn,letterpaper]{article}

\usepackage{cvpr}
\usepackage{times}
\usepackage{epsfig}
\usepackage{graphicx}
\usepackage{amsmath}
\usepackage{amssymb}

\usepackage{nicefrac}       
\usepackage{microtype}      

\usepackage{multirow}
\usepackage{verbatim}
\usepackage{color}
\usepackage{float}
\usepackage{enumitem}
\usepackage{booktabs}
\usepackage{tabulary,multirow,overpic,xcolor}

\usepackage[caption=false]{subfig}
\usepackage{pifont}
\usepackage{epstopdf}
\usepackage{algorithm}
\usepackage{algorithmicx}
\usepackage{algpseudocode}
\usepackage{enumerate}

\DeclareMathOperator*{\argmax}{argmax}

\usepackage{bm}
\usepackage{t1enc}
\usepackage{colortbl}
\usepackage{cite}

\usepackage[british,UKenglish,USenglish,english,american]{babel}

\newcommand{\figref}[1]{Fig.~\ref{#1}}
\newcommand{\tblref}[1]{Table~\ref{#1}}
\newcommand{\sref}[1]{Sec.~\ref{#1}}

\usepackage{soul}
\usepackage{soul}
\definecolor{carmine}{rgb}{0.59, 0.0, 0.09}

\newcommand{\app}{\raise.17ex\hbox{$\scriptstyle\sim$}}

\def\x{$\times$}

\newcolumntype{x}[1]{>{\centering\arraybackslash}p{#1pt}}

\newlength\savewidth\newcommand\shline{\noalign{\global\savewidth\arrayrulewidth
		\global\arrayrulewidth 1pt}\hline\noalign{\global\arrayrulewidth\savewidth}}
\newcommand{\tablestyle}[2]{\setlength{\tabcolsep}{#1}\renewcommand{\arraystretch}{#2}\centering\footnotesize}

\makeatletter\renewcommand\paragraph{\@startsection{paragraph}{4}{\z@}
	{.5em \@plus1ex \@minus.2ex}{-.5em}{\normalfont\normalsize\bfseries}}\makeatother

\definecolor{citecolor}{RGB}{34,139,34}
\usepackage[pagebackref=true,breaklinks=true,letterpaper=true,colorlinks,
citecolor=citecolor,bookmarks=false]{hyperref}

\cvprfinalcopy 

\renewcommand{\omega}{\alpha}
\renewcommand{\phi}{\beta}

\def\cvprPaperID{***} 

\begin{document}
				\title{X3D: Expanding Architectures for Efficient Video Recognition \vspace{-.8em}}
\author{
	Christoph Feichtenhofer\vspace{.8em}\\
	Facebook AI Research (FAIR)\vspace{.9em}
}

	\maketitle
	
	\definecolor{fastcolor}{RGB}{100,178,100}
	\definecolor{slowcolor}{RGB}{120,120,243}
	\definecolor{expandcolor}{RGB}{244,157,78}
	\definecolor{clipscolor}{HTML}{0071bc}
	
	\newcommand{\fastcolor}[1]{\textcolor{fastcolor}{#1}}
	\newcommand{\fastcolorC}[1]{\textcolor{orange}{#1}}
	\newcommand{\slowcolor}[1]{\textcolor{slowcolor}{#1}}
	\newcommand{\expandcolor}[1]{\textcolor{expandcolor}{#1}}
	\newcommand{\clipscolor}[1]{\textcolor{clipscolor}{#1}}
		
	\newcommand{\slow}{\slowcolor{Slow }}
	\newcommand{\fast}{\fastcolor{Fast }}
	
	\definecolor{predictioncolor}{RGB}{0,255,0}
	\definecolor{labelcolor}{RGB}{255,0,0}
	\newcommand{\predictioncolor}[1]{\textcolor{predictioncolor}{#1}}
	\newcommand{\labelcolor}[1]{\textcolor{labelcolor}{#1}}
	
	\newcommand{\pred}{\predictioncolor{\textbf{Predictions}: }}
	\newcommand{\gt}{\labelcolor{\textbf{Labels}: }}

	\definecolor{demphcolor}{RGB}{144,144,144}
	\newcommand{\demph}[1]{\textcolor{demphcolor}{#1}}
	
	\definecolor{xycolor}{RGB}{60, 120, 216}
		\newcommand{\xycolor}[1]{\textcolor{xycolor}{#1}}
	\definecolor{wcolor}{RGB}{103, 78, 167}
			\newcommand{\wcolor}[1]{\textcolor{wcolor}{#1}}
	\definecolor{dcolor}{RGB}{166, 77,21}
			\newcommand{\dcolor}[1]{\textcolor{dcolor}{#1}}
	\definecolor{gcolor}{RGB}{204, 102, 153}
			\newcommand{\gcolor}[1]{\textcolor{gcolor}{#1}}
	\definecolor{tcolor}{RGB}{80, 200, 180}
		\newcommand{\tcolor}[1]{\textcolor{tcolor}{#1}}
		\definecolor{eicolor}{RGB}{153, 51, 102}
	\newcommand{\eicolor}[1]{\textcolor{eicolor}{#1}}
	
	\def\gab{\textcolor{eicolor}{ $\bm{\gamma_{b}}$}}
	\def\gaw{\textcolor{wcolor}{ $\bm{\gamma_w$}}}
	\def\gag{\textcolor{gcolor}{ $\bm{\gamma_{g}$}}}
	\def\gax{\textcolor{xycolor}{$\bm{\gamma_x$}}}
	\def\gay{\textcolor{xycolor}{$\bm{\gamma_y$}}}
	\def\gaxy{\textcolor{xycolor}{$\bm{\gamma_{s}$}}}
	\def\gat{\textcolor{tcolor}{$\bm{\gamma_t$}}}
	\def\gatau{\textcolor{fastcolor}{$\bm{\gamma_\tau$}}}
	\def\gabeta{\textcolor{orange}{$\bm{\gamma_\beta$}}}
	\def\gaalpha{\textcolor{fastcolor}{$\bm{\gamma_\alpha$}}}
	\def\gad{\textcolor{dcolor}{$\bm{\gamma_d$}}}
	
			\vspace*{-0.8em}
\begin{abstract}
	\vspace{-.2em}
	This paper presents X3D, a family of efficient video networks that progressively expand a tiny 2D image classification architecture along multiple network axes, in space, time, width and depth. Inspired by feature selection methods in machine learning, a simple stepwise network expansion approach is employed that  expands a single axis in each step, such that good accuracy to complexity trade-off is achieved. To expand X3D to a specific target complexity, we perform progressive forward expansion followed by backward contraction. X3D achieves state-of-the-art performance while requiring 4.8\x~and 5.5\x~fewer multiply-adds and parameters for similar accuracy as previous work. Our most surprising finding is that networks with high spatiotemporal resolution can perform well, while being extremely light in terms of network width and  parameters. We report competitive accuracy at unprecedented efficiency on video classification and detection benchmarks. Code will be available at: \url{https://github.com/facebookresearch/SlowFast}. 
	
\end{abstract}
\vspace{-5pt}	
\section{Introduction}

\label{sec:introduction}

Neural networks for video recognition have been largely driven by \emph{expanding} 2D image architectures \cite{Krizhevsky2012,Simonyan2015,Szegedy2015,He2016} into spacetime.
Naturally, these expansions often happen along the temporal axis, involving extending the network inputs, features, and/or filter kernels into spacetime (\eg \cite{Karpathy2014,Donahue2015,Ng2015,Tran2015,Feichtenhofer2016a,Carreira2017}); other design decisions---including depth (number of layers), width (number of channels), and spatial sizes---however, are typically inherited from 2D image architectures. While expanding along the temporal axis (while keeping other design properties) generally increases accuracy, it can be sub-optimal if one takes into account the computation/accuracy \emph{trade-off}---a consideration of central importance in applications.

In part because of the direct extension of 2D models to 3D, video recognition architectures are computationally heavy. In comparison to image recognition, typical video models are significantly more compute-demanding, \eg an image ResNet \cite{He2016} can use around 27\x~fewer multiply-add operations than a temporally extended video variant \cite{Wang2018}.

This paper focuses on the low-computation regime in terms of computation/accuracy {trade-off} for video recognition. We base our design upon the ``mobile-regime" models \cite{Howard2017,Sandler2018,Howard2019} developed for image recognition. Our core idea is that while expanding a small model along the temporal axis can increase accuracy, the computation/accuracy trade-off may not always be best compared with expanding \emph{other axes}, especially in the low-computation regime where accuracy can increase quickly along different axes. 
\begin{figure}[t]
	\centering
	\vspace{-1em}
	\includegraphics[width=1\linewidth]{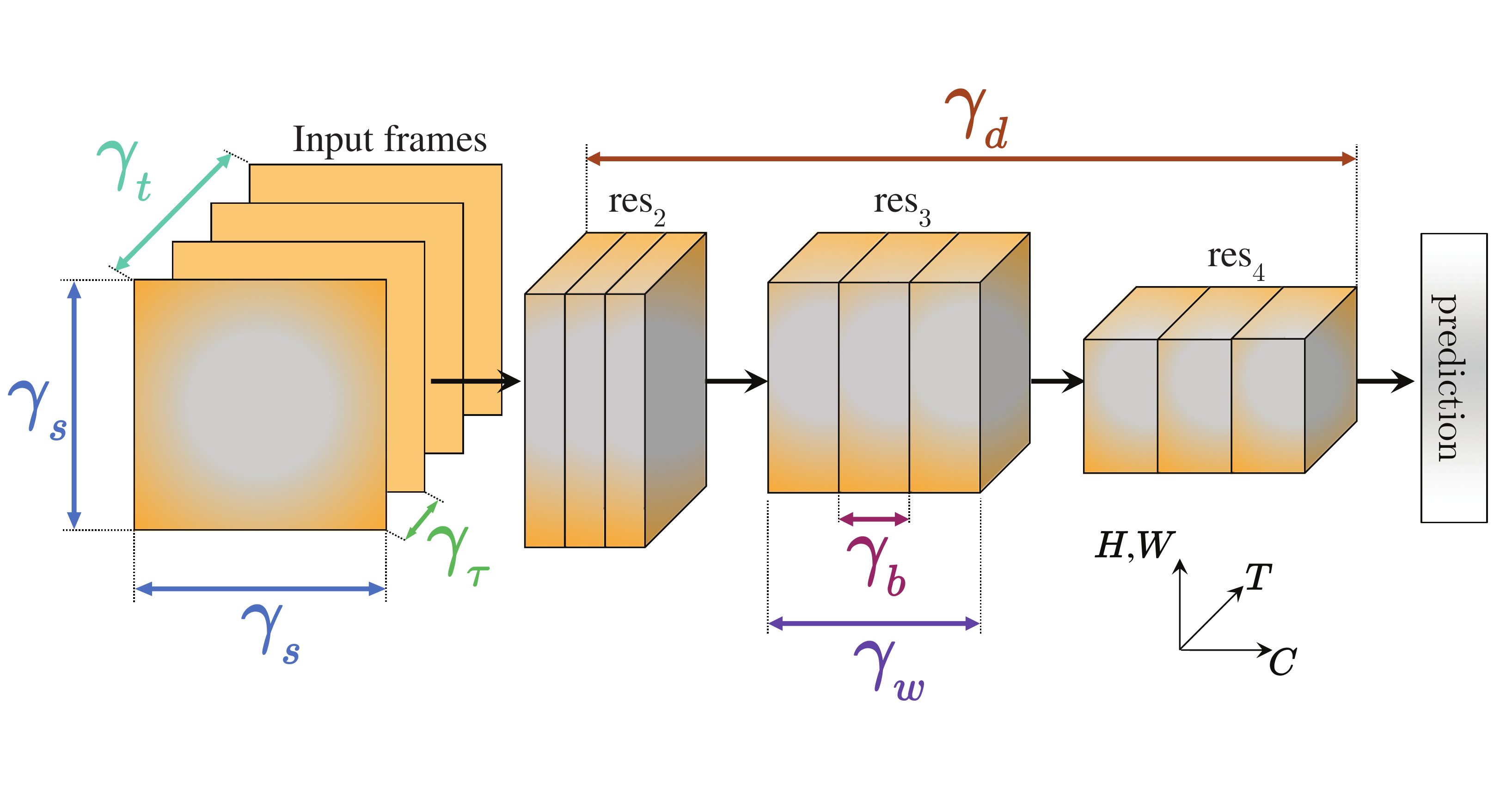}
		\vspace{-30pt}

	\caption{{\textbf{{X3D}}} networks progressively {\color{expandcolor}{expand}} a 	{\color{gray}{2D network}}  across the following axes: Temporal duration \gat,  frame rate \gatau,  spatial resolution \gaxy, width \gaw,  bottleneck width \gab, and depth \gad.
	}
	\label{fig:teaser}
\end{figure}

In this paper, we progressively ``expand" a tiny base 2D image architecture into a {spatiotemporal one} by expanding \emph{multiple} possible axes shown in  \figref{fig:teaser}.
The candidate axes are temporal duration  \gat, frame rate \gatau, spatial resolution \gaxy, network width \gaw, bottleneck width \gab, and depth \gad. The resulting architecture is referred as \textbf{X3D} (Expand 3D) for \expandcolor{expanding} from the {\color{gray}{2D space}} into {3D spacetime} domain.

The 2D base architecture is driven by the MobileNet \cite{Howard2017,Sandler2018,Howard2019} core \emph{concept} of channel-wise\footnote{Also referred as ``depth-wise". We use the term ``channel-wise" to avoid confusions with the network depth, which is also an axis we consider.} separable convolutions, but is made tiny by having over 10\x~fewer multiply-add operations than mobile image models. Our expansion then progressively increases the computation (\eg, by 2$\times$) by expanding only \emph{one axis at a time}, train  and validate the resultant architecture, and select the axis that achieves the best computation/accuracy trade-off. The process is repeated until the architecture reaches a desired computational budget. This  can be interpreted as a form of \emph{coordinate descent} \cite{wright2015coordinate} in the hyper-parameter space defined by those axes.

\vspace{-1.2em}

Our progressive network expansion approach is inspired by the history of image ConvNet design where popular architectures have arisen by expansions across depth, \cite{Krizhevsky2012,Zeiler2012,Chatfield2014a,Simonyan2015,Szegedy2015,He2016},  resolution \cite{Szegedy2016,huang2018gpipe,tan2019efficientnet} or width \cite{Zagoruyko2016,Xie2017},  and classical feature selection methods \cite{john1994irrelevant,kohavi1997wrappers,guyon2003introduction} in machine learning. In the latter, progressive feature selection methods \cite{kohavi1997wrappers,guyon2003introduction} start with either a set of minimum features and aim to find relevant features to improve in a greedy fashion by including  (\textit{forward selection}) a single feature in each step, or  start with a full set of features and aim to find irrelevant ones that are excluded by repeatedly deleting the feature that reduces performance the least (\textit{backward elimination}).
	
To compare to previous research, we use Kinetics-400 \cite{Kay2017}, {Kinetics-600} \cite{Carreira2018}, Charades \cite{Sigurdsson2016}  and AVA \cite{Gu2018}.
For systematic studies, we classify our models into different levels of complexity for small, medium and large models.

Overall, our expansion produces a sequence of spatiotemporal architectures,  covering a wide range of computation/accuracy trade-offs. They can be used under different computational budgets that are application-dependent in practice. For example, across different computation and accuracy regimes X3D  performs favorably to state-of-the-art while requiring 4.8\x~and 5.5\x~fewer multiply-adds and parameters for similar accuracy as previous work. Further, expansion is simple and cheap \eg our low-compute model is completed after only training 30 \textit{tiny} models that \textit{accumulatively} require over 25\x~fewer multiply-add operations for training than one large state-of-the-art network \cite{Wang2018,Feichtenhofer2019,Wu2019}.
	
Conceptually, our most surprising finding is that \textit{very thin} video architectures that are created by expanding spatio-temporal resolution perform well, while being  light in terms of network width and  parameters. X3D networks have lower width than image-design \cite{Simonyan2015,Szegedy2015,He2016} based video models, making X3D similar to the high-resolution Fast pathway \cite{Feichtenhofer2019} which has been designed in such fashion. 
 We hope these advances will facilitate future research and applications. 
 

	\section{Related Work}
\label{sec:related_work}

\paragraph{Spatiotemporal (3D)  networks.} 

Video recognition architectures are favorably designed by extending image classification networks with a temporal dimension, and preserving the spatial properties. These extensions include direct transformation of 2D models  \cite{Krizhevsky2012,Simonyan2015,Szegedy2015,He2016}  such as ResNet or Inception to 3D  \cite{Taylor2010,Tran2015,Carreira2017,Qiu2017,hara2018can,Xie2018}, adding RNNs on top of 2D CNNs \cite{Donahue2015, Ng2015, Yosinski2015,Sun2017,Li2018,Li2018a}, or extending 2D models with an optical flow stream that is processed by an identical 2D network ~\cite{Simonyan2014,Wang2016a,Feichtenhofer2016, Carreira2017} .  While starting with a 2D image based model and converting it to a spatiotemporal equivalent by inflating filters \cite{Feichtenhofer2016a,Carreira2017} allows pretraining on image classification tasks, it makes video architectures inherently biased towards their image-based counterparts. 

The SlowFast \cite{Feichtenhofer2019} architecture has explored the resolution trade-off across several axes, different temporal, spatial, and channel resolution in the Slow and Fast pathway. Interestingly the Fast pathway can be very thin and therefore only adds a small computational overhead; however, performs low in isolation. Further, these explorations were performed with the architecture of the computationally heavy Slow pathway held constant to a  temporal extension of an image classification design \cite{He2016}. In relation to this previous effort, our work investigates whether the heavy Slow pathway is required, or if a lightweight network can be made competitive.

\paragraph{Efficient 2D networks.}
Computation-efficient architectures have been extensively developed for the image classification task, with \mbox{MobileNetV1\&2} \cite{Howard2017,Sandler2018} and ShuffleNet \cite{Zhang2018} exploring channel-wise separable convolutions and expanded bottlenecks. Several methods for neural architecture search in this setting have been proposed, also adding Squeeze-Excitation (SE) \cite{hu2018squeeze} attention blocks to the design space in \cite{tan2019mnasnet} and more recently, MobileNetV3 \cite{Howard2019} Swish non-linearities  \cite{ramachandran2017searching}.  MobileNets \cite{Howard2017,Sandler2018,tan2019mnasnet} were scaled up and down by using a multiplier for width and input size (resolution). Recently, MnasNet \cite{tan2019mnasnet} is used to apply liner scaling factors to spatial, width and depth axes for creating a set of EfficientNets \cite{tan2019efficientnet} for image classification.

Our expansion is related to this, but requires fewer samples and handles more axes as we only train a single model for each axis in each step, while \cite{tan2019efficientnet} performs a grid-search on the initial regime which requires $k^d$ models to be trained where $k$ is the gridsize and $d$ the number of axes.
Moreover, the model used for this search, MnasNet was found by sampling around 8000 models \cite{tan2019mnasnet}. For video, this is prohibitive as datasets can have  orders of magnitude more images than image classification \eg the largest version of Kinetics \cite{Carreira2019} has $\approx$195M frames, 162.5$\times$ more images than ImageNet.
By contrast, our approach only requires to train 6 models, one for each expansion axis, until a desired complexity is reached, \eg for 5 steps, it requires 30 models to be trained. 

\paragraph{Efficient 3D networks.} 
Several innovative architectures for efficient video classification have been proposed, \eg \cite{Carreira_2018_ECCV,Xie2018,kopuklu2019resource,Tran2019,chen2018multi,Zolfaghari2018,lee2018motion,Bilen2016, fernando2015modeling,Wu2018,Sun2015,fan2018end,Sun_2018_CVPR,zhu2017hidden,Piergiovanni_2019_CVPR,Diba2018,Wang2018a,Luo_2019_ICCV,hussein2019timeception,zhu2019faster,wang2019video}. 
Channel-wise separable convolution as a key building block for efficient 2D ConvNets \cite{Howard2017,Sandler2018,Howard2019,Zhang2018,tan2019efficientnet} has been explored for video classification in \cite{kopuklu2019resource,Tran2019}, where 2D architectures are extended to their 3D counterparts, \eg ShuffleNet and MobileNet in \cite{kopuklu2019resource}, or ResNet in  \cite{Tran2019} by using a 3\x3\x3 channel-wise separable convolution in the bottleneck of a residual stage. Earlier, \cite{chen2018multi} adopt 2D ResNets and MobileNets from ImageNet and sparsifies connections inside each residual block similar to separable or group convolution. A temporal shift module (TSM) is introduced in \cite{lin2018temporal} that extends a ResNet to capture temporal information using memory shifting operations. There is also active research on adaptive frame sampling techniques, \eg \cite{yeung2016end,su2016leaving,alwassel2018action,wu2019adaframe,Wu_2019_ICCV,korbar2019scsampler}, which we think can be complementary to our approach.

\vspace{15pt}
In relation to most of these works, our approach does not assume a fixed inherited design from 2D networks, but expands a tiny architecture across several axes in space, time, channels and depth to achieve a good efficiency trade-off.

\section{X3D Networks}  \label{sec:x3d}

Image classification architectures have gone through an evolution of architecture design with progressively \textit{expanding} existing models along network depth \cite{Krizhevsky2012,Zeiler2012,Chatfield2014a,Simonyan2015,Szegedy2015,He2016}, input resolution \cite{Szegedy2016,huang2018gpipe,tan2019efficientnet} or channel width \cite{Zagoruyko2016,Xie2017}.
Similar progress can be observed for the mobile image classification domain  where \textit{contracting} modifications (shallower networks, lower resolution, thinner layers, separable convolution \cite{Iandola2016, Howard2017,Sandler2018,Zhang2018,Howard2019}) allowed operating at lower computational budget.
Given this history in image ConvNet design, a similar progress has not been observed for video architectures as these were customarily based on direct temporal extensions of image models. However, is single expansion of a \textit{fixed} 2D architecture to 3D ideal, or is it better to \textit{expand} or \textit{contract} along different axes?

For video classification the temporal dimension exposes an additional dilemma, increasing the number of possibilities but also requiring it to be dealt differently than the spatial dimensions \cite{Simonyan2014,Tran2018, Feichtenhofer2019}. We are especially interested in the trade-off between different axes, more concretely:

\begin{itemize}
\setlength\itemsep{.2em}
	
	\item What is the best temporal sampling strategy for 3D networks? Is a \textit{long} input duration and sparser sampling preferred over \textit{faster} sampling of short duration clips? 
	\item Do we require \textit{finer} spatial resolution? Previous works have used lower resolution for video classification \cite{Karpathy2014, Tran2015,Tran2018} to increase efficiency. Also, videos typically come at \textit{coarser} spatial resolution than Internet images; therefore, is there a maximum spatial resolution at which performance saturates?  
	\item Is it better to have a network with high frame-rate but \textit{thinner} channel resolution, or to slowly process video with a \textit{wider} model? \textit{E.g.~}should the network have heavier layers as typical image classification models (and the Slow pathway  \cite{Feichtenhofer2019}) or rather lighter layers with lower width (as the Fast pathway  \cite{Feichtenhofer2019}). Or is there a better trade-off, possibly between these extremes?
	\item When increasing the network width, is it better to globally expand the network width in the ResNet block design \cite{He2016} or to expand the {inner} (``\textit{bottleneck}'') width, as is common in mobile image classification networks using channel-wise separable convolutions \cite{Sandler2018,Zhang2018}?
	\item Should going \textit{deeper} be performed with expanding input resolution in order to keep the receptive field size large enough and its growth rate roughly constant, or is it better to expand into different axes? Does this hold for both the spatial and temporal dimension? 
\end{itemize}

This section first introduces the basis X2D architecture in \sref{sec:basis} which is expanded with operations defined in \sref{sec:x_ops} by using the progressive approach in \sref{sec:x_alg}.  

\newcommand{\blocks}[3]{\multirow{3}{*}{\(\left[\begin{array}{c}\text{1$\times$1$^\text{2}$, #2}\\[-.1em] \text{1$\times$3$^\text{2}$, #2}\\[-.1em] \text{1$\times$1$^\text{2}$, #1}\end{array}\right]\)$\times$#3}
}
\newcommand{\blocket}[4]{\multirow{3}{*}{\(\left[\begin{array}{c}\text{1$\times$1$^\text{2}$, #1}\\[-.1em] \text{$3$$\times$3$^\text{2}$, #2}\\[-.1em] \text{1$\times$1$^\text{2}$, #3}\end{array}\right]\)$\times$#4}
}
\newcommand{\blockt}[3]{\multirow{3}{*}{\(\left[\begin{array}{c}\text{\underline{3$\times$1$^\text{2}$}, #2}\\[-.1em] \text{1$\times$3$^\text{2}$, #2}\\[-.1em] \text{1$\times$1$^\text{2}$, #1}\end{array}\right]\)$\times$#3}
}
\newcommand{\outsizes}[7]{\multirow{#7}{*}{\(\begin{array}{c} \text{\emph{Slow}}: \text{#1$\times$#2$^\text{2}$}\\[-.1em] \text{\emph{Fast}}: \text{#4$\times$#5$^\text{2}$}\end{array}\)}
}

\newcommand{\outsizesRaw}[4]{\multirow{#4}{*}{\(\begin{array}{c}  \text{#1$\times$#2\x #3}\\[-.1em]  \end{array}\)}}
\newcommand{\outsizesGamma}[4]{\multirow{#4}{*}{\(\begin{array}{c}  \text{#1\gat\x (#2\gaxy)$^2$}\\[-.1em]  \end{array}\)}}

\newcommand{\outsizesSF}[5]{\multirow{#5}{*}{\(\begin{array}{cc} \text{\emph{Slow}}:& \text{#1$\times$ #2$^\text{2}$}\\[-.1em] \text{\emph{Fast}}:& \text{#3$\times$#4$^\text{2}$}\end{array}\)}}

\begin{table}[t]
	\scriptsize
	\centering
	\resizebox{\columnwidth}{!}{
		\tablestyle{1pt}{1.08}
		\begin{tabular}{c|c|c}
			stage & filters & output sizes $T$\x$S^2$ \\
			\shline
			\multirow{1}{*}{data layer} & \multirow{1}{*}{stride \gatau, 1$^\text{2}$}  &  \outsizesGamma{{1}}{112}{112}{1}   \\
			\hline
			\multirow{1}{*}{conv$_1$} & \multirow{1}{*}{1\x3$^\text{2}$, 3\x1, {24\gaw}}  &  \outsizesGamma{{1}}{56}{56}{1}    \\
			\hline
			\multirow{3}{*}{res$_2$}  & \blocket{{24\gab\gaw}}{{24\gab\gaw}}{{24\gaw}}{\gad} & \outsizesGamma{{1}}{28}{28}{3}  \\
			&  & \\
			&  & \\
			\hline
			\multirow{3}{*}{res$_3$}  & \blocket{{48\gab\gaw}}{{48\gab\gaw}}{{48\gaw}}{2\gad} & \outsizesGamma{{1}}{14}{14}{3}  \\
			&  & \\
			&  & \\
			\hline
			\multirow{3}{*}{res$_4$}  & \blocket{{96\gab\gaw}}{{96\gab\gaw}}{{96\gaw}}{5\gad} & \outsizesGamma{{1}}{7}{7}{3}  \\
			&  & \\
			&  & \\
			\hline
			\multirow{3}{*}{res$_5$}  & \blocket{{192\gab\gaw}}{{192\gab\gaw}}{{192\gaw}}{3\gad} & \outsizesGamma{{1}}{4}{4}{3}  \\
			&  & \\
			&  & \\
			\hline
			\multirow{1}{*}{conv$_5$} & \multirow{1}{*}{1\x1$^\text{2}$, {192\gab\gaw}}   &  \outsizesGamma{{1}}{4}{4}{1}    \\
			\multirow{1}{*}{pool$_5$} &\outsizesGamma{{1}}{4}{4}{1}   & 1\x1\x1    \\
			\multirow{1}{*}{fc$_1$} & \multirow{1}{*}{1\x1$^\text{2}$, {2048}}    & 1\x1\x1    \\
			\multicolumn{1}{c|}{ fc$_2$}  & \multirow{1}{*}{1\x1$^\text{2}$, { \#classes}} & 1\x1\x1  \\
	\end{tabular}}
	
	\vspace{.1em}
	\caption{\textbf{X3D architecture}. The dimensions of kernels are denoted by $\{$$T$\x $S^2$, $C$$\}$ for temporal, spatial, and channel sizes.
		Strides are denoted as $\{$temporal stride, spatial stride$^2$$\}$.
		This network is expanded using factors $\{$\gatau, \gat, \gaxy, \gaw, \gab,  \gad$ \}$ to form \textbf{X3D}. Without expansion (all factors equal to one), this model is referred to as \textbf{X2D}, having 20.67M FLOPS and 1.63M parameters.
	}
	\label{tab:arch}
	\vspace{-.8em}
\end{table}

\subsection{Basis instantiation}  \label{sec:basis}
We  begin by describing the instantiation of the basis network architecture, {X2D}, that serves as baseline to be expanded into spacetime. The basis network instantiation follows a ResNet \cite{He2016} structure and the Fast pathway design of SlowFast networks \cite{Feichtenhofer2019} with degenerated (single frame) temporal input. {X2D} is specified in Table~\ref{tab:arch}, if all expansion factors $\{$\gatau, \gat, \gaxy, \gaw, \gab,  \gad$ \}$ are set to 1.

We denote spatiotemporal size by $T$\x $S^2$ where $T$ is the temporal length and $S$ is the height and width of a square spatial crop. The X2D architecture is described next.

\paragraph{Network resolution and channel capacity.} The model takes as input a raw video clip that is sampled with frame-rate 1$/$\gatau~in the data layer stage. The basis architecture only takes a single frame of size $T$\x $S^2$$=$1\x112$^2$  as input and therefore can be seen as an image classification network. The width of the individual layers is oriented at the Fast pathway design in \cite{Feichtenhofer2019} with the first stage, conv$_1$, filters the 3 RGB input channels and produces 24 output features. This width is increased by a factor of 2 after every spatial sub-sampling with a stride $=1, 2^2$ at each deeper stage from res$_2$ to res$_5$. Spatial sub-sampling is performed by the center (``bottleneck'') filter of the first res-block of each stage.

Similar to the SlowFast pathways \cite{Feichtenhofer2019}, the model preserves the temporal input resolution for all features throughout the network hierarchy. There is no temporal downsampling layer (neither temporal pooling nor time-strided convolutions) throughout the network, up to the global pooling layer before classification. Thus, the activations tensors contain all frames along the temporal dimension, maintaining full temporal frequency in all features.

\paragraph{Network stages.}
X2D consists of a stage-level and bottleneck design that is inspired by recent 2D mobile image classification networks \cite{Howard2017,Sandler2018,Howard2019,Zhang2018} which employ channel-wise separable convolution that are a key building block for efficient ConvNet models. 
 We adopt stages that follow MobileNet \cite{Sandler2018,Howard2019} design by extending every spatial 3\x3 convolution in the bottleneck block to a 3\x3\x3 (\ie 3\x3$^\text{2}$) spatiotemporal convolution  which has also been explored for video classification in \cite{kopuklu2019resource,Tran2019}. Further, the 3\x1 temporal convolution in the first conv$_1$ stage is channel-wise.

\paragraph{Discussion.}
X2D can be interpreted as a Slow pathway since it only uses a single frame as input, while the network width is similar to the Fast pathway in \cite{Feichtenhofer2019} which is much lighter than typical 3D ConvNets (\eg, \cite{Feichtenhofer2016a,Tran2015,Carreira2017,Wang2018,Feichtenhofer2019}) that follow an ImageNet design. Concretely, it only requires 20.67M FLOPs which amounts to only 0.0097\% of a recent state-of-the-art SlowFast network \cite{Feichtenhofer2019}.

As shown in Table~\ref{tab:arch} and \figref{fig:teaser}, \textbf{ X2D} is expanded across 6 axes, $\{$\gatau, \gat, \gaxy, \gaw, \gab,  \gad$ \}$, described next.

\subsection{Expansion operations}  \label{sec:x_ops}

We define a basic set of expansion operations that are used for sequentially expanding \textbf{X2D} from a tiny spatial network to \textbf{X3D}, a spatiotemporal network, by performing the following operations on temporal, spatial, width and depth dimensions. 
\begin{itemize}
	\item \textbf{\textit{\fastcolor{X-Fast}}} expands the temporal activation size, \gat,~by increasing the frame-rate, 1$/$\gatau, and therefore temporal resolution,~while holding the clip duration constant. 
	\item \textbf{\textit{\tcolor{X-Temporal}}} expands the temporal size, \gat,~by sampling a longer temporal clip and increasing the frame-rate 1$/$\gatau, to expand both duration \textit{and} temporal resolution. 
	\item \textbf{\textit{\xycolor{X-Spatial}}} expands the spatial resolution, \gaxy,~by increasing the spatial sampling resolution of the input video. 
	\item \textbf{\textit{\dcolor{X-Depth}}} expands the depth of the network by increasing the number of layers per residual stage by \gad~times.
	\item \textbf{\textit{\wcolor{X-Width}}} uniformly expands the channel number for all layers by a global width expansion factor \gaw. 
	\item \textbf{\textit{\eicolor{X-Bottleneck}}} expands the inner channel width, \gab, of the center convolutional filter in each residual block.
\end{itemize}

\subsection{Progressive Network Expansion}  \label{sec:x_alg}

We employ a simple progressive algorithm for network expansion, similar to forward and backward algorithms for feature selection \cite{john1994irrelevant,kohavi1997wrappers, jain1997feature, guyon2003introduction}. Initially we start with X2D, the basis model instantiation with a set of unit expanding factors $\mathcal{X}_0$ of cardinality $a$. We use $a=6$ factors, $\mathcal{X}=$$\{$\gatau, \gat, \gaxy, \gaw, \gab, \gad$ \}$, but other axes are possible. 
\paragraph{Forward expansion.}
The network expansion criterion function, which measures the goodness for the current expansion factors $\mathcal{X}$, is represented as $J(\mathcal{X})$. Higher scores of this measure represent better expanding factors, while lower scores would represent worse. In our experiments, this corresponds to the accuracy of a model expanded by $\mathcal{X}$. Furthermore, let $C(\mathcal{X})$ be a complexity criterion function that measures the cost of the current expanding factors $\mathcal{X}$. In our experiments, $C$ is set to the floating point operations of the underlying network instantiation expanded by $\mathcal{X}$, but other measures such as runtime, parameters, or memory are possible. Then, the network expansion tries to find expansion factors $\mathcal{X}$ with the best trade-off
$
\mathcal{X} = \argmax_{\mathcal{Z}, C(\mathcal{Z}) = c} = J(\mathcal{Z}) \label{eq:expansion}
$
where $\mathcal{Z}$ are the possible expansion factors to be explored and $c$ is the target complexity. In our case we perform expansion that only changes a \textit{single one} of the $a$ expansion factors while holding the others constant; therefore there are only $a$ different subsets of $\mathcal{Z}$ to evaluate, where each of them alters in only one dimension from $\mathcal{X}$. The expansion with the best computation/accuracy trade-off is kept for the next step. This is a form of \emph{coordinate descent} \cite{wright2015coordinate} in the hyper-parameter space defined by those axes.

The expansion is performed in a progressive manner with an expansion-rate $\hat{c}$ that corresponds to the stepsize at which the model complexity $c$ is increased in each expansion step. We use a multiplicative increase of $\hat{c}\approx2$ of the model complexity in each step that corresponds to the complexity-increase for doubling the number of frames of the model. 
The stepwise expansion is therefore simple and efficient as it only requires to train a few models until a target complexity is reached, since we \textit{exponentially} increase the complexity.  Details on the expansion are in \S\ref{sec:kineticsapp}.  
\paragraph{Backward contraction.} Since the forward expansion only produces models in discrete steps, we perform a backward contraction step to meet a desired target complexity, if the target is exceeded by the forward expansion steps. This contraction is implemented as a simple reduction of the last expansion, such that it matches the target. For example, if the last step has increased the frame-rate by a factor of two, the backward contraction will reduce the frame-rate by a factor $<$ 2 to roughtly match the desired target complexity. 

	\section{Experiments: Action Classification}\label{sec:kinetics}

	\paragraph{Datasets.} We perform our expansion on Kinetics-400 \cite{Kay2017} (K400)  with $\app$240k training, 20k validation and 35k testing
	videos in 400 human action categories. 
	We report top-1 and top-5 classification accuracy (\%). 
	As in previous work, we train and report ablations on the {train} and {val} sets. We also report results on {test} set as the labels have been made available \cite{Carreira2018}.
		We report the computational cost (in FLOPs) of a single, spatially center-cropped clip.\footnote{We use single-clip, center-crop FLOPs as a basic \emph{unit} of computational cost. Inference-time computational cost is roughly proportional to this, \emph{if} a fixed number of clips and crops is used, as is for our all models.}
	
	\paragraph{Training.} All models are trained \emph{from random initialization} (``\emph{from scratch}'') on Kinetics, \emph{without} using ImageNet \cite{Deng2009} or other pre-training. Our training recipe follows \cite{Feichtenhofer2019}. 
	All implementation details and dataset specifics are in \S\ref{sec:kineticsapp}. 
	
	For the temporal domain, we randomly sample a clip from the full-length video, and the input to the network are  \gat~frames with a temporal stride of \gatau; for the spatial domain, we randomly crop 112\gaxy\x112\gaxy~pixels from a video, or its horizontal flip, with a shorter side randomly sampled in [128\gaxy, 160\gaxy] pixels which is a linearly scaled version of the augmentation used  in \cite{Simonyan2015, Wang2018, Feichtenhofer2019}.

\paragraph{Inference.}  To be comparable with previous work and evaluate accuracy/complexity trade-offs we apply two testing strategies: \textit{(i) $K$-Center}: Temporally, uniformly samples $K$ clips  (\eg $K$$=$10)  from a video and spatially scales the shorter spatial side to 128\gaxy~pixels and takes a \gat\x112\gaxy\x112\gaxy~center crop, comparable to ~\cite{korbar2019scsampler,Wu2019,Tran2019,lin2018temporal}.
\textit{(ii) $K$-LeftCenterRight} is the same as above temporally, but takes 3 crops of \gat\x128\gaxy\x128\gaxy~to cover the longer spatial axis, as an approximation of fully-convolutional testing, following \cite{Wang2018,Feichtenhofer2019}.
We average the softmax scores for all individual predictions.

  	\begin{figure}[t]
  	\vspace{-10pt}
  	\centering
  \hspace*{-15pt}	\includegraphics[width=1.10\linewidth]{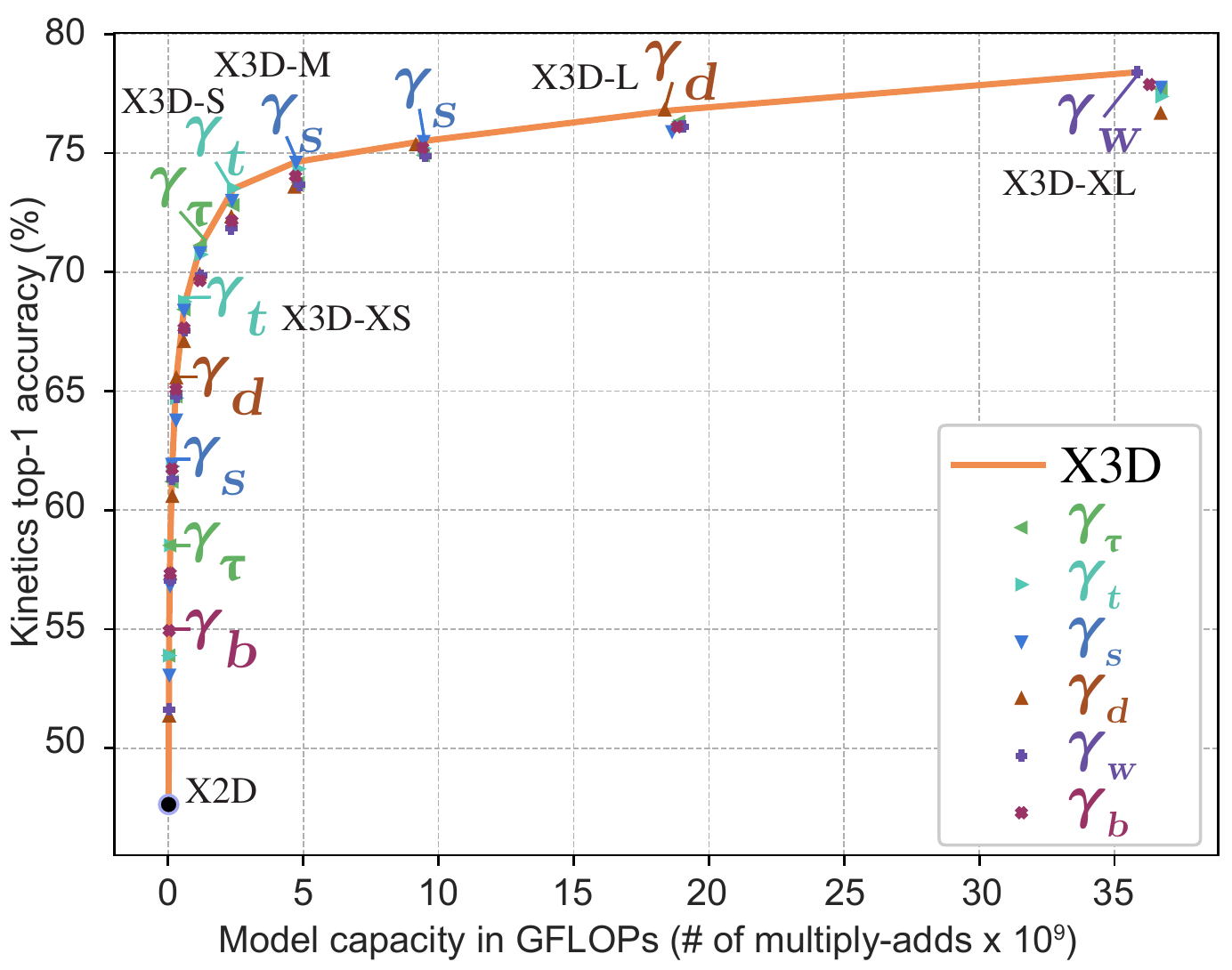} 
  	\caption{\textbf{Progressive network expansion of X3D.} The X2D base model is expanded 1$^\text{st}$ across bottleneck width (\gab), 2$^\text{nd}$  temporal resolution (\gatau), 3$^\text{rd}$ spatial resolution (\gaxy), 4$^\text{th}$  depth (\gad), 5$^\text{th}$  duration (\gat), \etc 
  		The majority of models are trained for small computation cost, making the expansion economical in practice. 
  	}
  	\label{fig:expansion_curve}
  	\vspace{-15pt}
  \end{figure}

\begin{table}[t]\centering  
	\vspace{-25pt}
	\tablestyle{1.8pt}{1.05}
	\begin{tabular}{l|x{28}|x{28}|r|x{28}|x{28}}
		\multirow{2}{*}{model} & \multirow{2}{*}{top-1} & \multirow{2}{*}{top-5} & 	\multicolumn{1}{c|}{{\multirow{1}{*}{regime}}}  & FLOPs & Params \\
		&&&	 FLOPs (G) &  (G)&(M) \\
		\shline
		\textbf{X3D-XS}  & 68.6 & 87.9  &  \emph{X-Small} $\leq$ 0.6  & 0.60 & 3.76 \\    
		\textbf{X3D-S} & 72.9 & 90.5 & \emph{Small} $\leq$ 2  &	 1.96 & 3.76 \\ 
		\textbf{X3D-M}  & 74.6 & 91.7 & \emph{Medium} $\leq$ 5 	 &4.73 & 3.76 \\  
		\textbf{X3D-L}  & 76.8 & 92.5  &  \emph{Large} $\leq$ 20 	& 18.37 & 6.08 \\  
		\textbf{X3D-XL} &   78.4 & 93.6  & \emph{X-Large} $\leq$ 40 	 &	35.84 & 11.0  \\ 
		\textbf{X3D-XXL} &   80.0 & 94.5  & \emph{XX-Large} $\leq$ 150	 &	143.5 & 20.3  \\  
	\end{tabular}
	\caption{Expanded instances on K400-val. 10-Center clip testing is used. We show top-1 and top-5 classification accuracy (\%), as well as computational complexity measured in GFLOPs (floating-point operations, in \# of multiply-adds \x $10^9$) for a  single clip input. Inference-time computational cost is proportional to 10\x~of this, as a fixed number of 10 of clips is used per video. 
	}
	\label{tab:expansion:x3d}
	\vspace{-15pt}
\end{table}

\subsection{Expanded networks}  \label{sec:xpanded_nets}
The accuracy/complexity trade-off curve for the expansion process on K400 is shown in \figref{fig:expansion_curve}. Expansion starts from X2D that produces 47.75\% top-1 accuracy (vertical axis) with 1.63M parameters 20.67M FLOPs per clip (horizontal axis), which is roughly doubled in each progressive  step. We use \textit{10-Center} clip testing as our default test setting for expansion, so the overall cost per video is \x10. We will ablate different number of testing clips in \sref{sec:ablations}.
The expansion  in \figref{fig:expansion_curve} provides several interesting observations:

\begin{table*}[t]
	\vspace{-3.5em}
	\scriptsize
	\centering
	\captionsetup[subfloat]{captionskip=2pt}
	\captionsetup[subffloat]{justification=centering}
	\subfloat[\textbf{X3D-S} with 1.96G FLOPs, 3.76M param, and  72.9\% top-1 accuracy using expansion of \gatau $=$ 6, \gat $=$ 13, \mbox{\gaxy $=$ $\sqrt{\text{2}}$,} \gaw $=$ 1, \gab $=$ 2.25, \gad $=$ 2.2.
	\label{tab:arch_s}]{
		\tablestyle{2pt}{1.05}
		\resizebox{0.31\linewidth}{!}{
			\tablestyle{1pt}{1.08}
			\begin{tabular}{c|c|c}
				stage & filters & output sizes $T$\x$H$\x$W$ \\
				\shline
				\multirow{1}{*}{data layer} & \multirow{1}{*}{stride 6, 1$^\text{2}$}  &  \outsizesRaw{13}{160}{160}{1}   \\
				\hline
				\multirow{1}{*}{conv$_1$} & \multirow{1}{*}{1\x3$^\text{2}$, 3\x1, {24}}  & \outsizesRaw{13}{80}{80}{1}   \\
				\hline
				\multirow{3}{*}{res$_2$}  & \blocket{{54}}{{54}}{{24}}{3} & \outsizesRaw{13}{40}{40}{3}  \\
				&  & \\
				&  & \\
				\hline
				\multirow{3}{*}{res$_3$}  & \blocket{{108}}{{108}}{{48}}{5} & \outsizesRaw{13}{20}{20}{3} \\
				&  & \\
				&  & \\
				\hline
				\multirow{3}{*}{res$_4$}  & \blocket{{216}}{{216}}{{96}}{11} & \outsizesRaw{13}{10}{10}{3} \\
				&  & \\
				&  & \\
				\hline
				\multirow{3}{*}{res$_5$}  & \blocket{{432}}{{432}}{{192}}{7} & \outsizesRaw{13}{5}{5}{3}  \\
				&  & \\
				&  & \\
				\hline
				\multirow{1}{*}{conv$_5$} & \multirow{1}{*}{1\x1$^\text{2}$, {432}}   & \outsizesRaw{13}{5}{5}{1}   \\
				\multirow{1}{*}{pool$_5$} &\outsizesRaw{{13}}{5}{5}{1}   & 1\x1\x1    \\
				\multirow{1}{*}{fc$_1$} & \multirow{1}{*}{1\x1$^\text{2}$, {2048}}    & 1\x1\x1    \\
				\multicolumn{1}{c|}{ fc$_2$}  & \multirow{1}{*}{1\x1$^\text{2}$, { \#classes}} & 1\x1\x1  \\
		\end{tabular}}
	}\hspace{5pt}
	\subfloat[\textbf{X3D-M}  with 4.73G FLOPs, 3.76M param, and 74.6\% top-1 accuracy using expansion of \mbox{\gatau $=$ 5}, \mbox{\gat $=$ 16}, \mbox{\gaxy $=$ 2,} \gaw $=$ 1, \gab $=$ 2.25, \gad $=$ 2.2.
	\label{tab:arch_m}]{
		\tablestyle{2pt}{1.05}
		\resizebox{0.31\linewidth}{!}{
			\tablestyle{1pt}{1.08}
			\begin{tabular}{c|c|c}
				stage & filters & output sizes $T$\x$H$\x$W$ \\
				\shline
				\multirow{1}{*}{data layer} & \multirow{1}{*}{stride \fastcolor{5}, 1$^\text{2}$}  &  \outsizesRaw{\tcolor{16}}{\xycolor{224}}{\xycolor{224}}{1}   \\
				\hline
				\multirow{1}{*}{conv$_1$} & \multirow{1}{*}{1\x3$^\text{2}$, 3\x1, {24}}  & \outsizesRaw{\tcolor{16}}{\xycolor{112}}{\xycolor{112}}{1}   \\
				\hline
				\multirow{3}{*}{res$_2$}  & \blocket{{54}}{{54}}{{24}}{3} & \outsizesRaw{\tcolor{16}}{\xycolor{56}}{\xycolor{56}}{3}  \\
				&  & \\
				&  & \\
				\hline
				\multirow{3}{*}{res$_3$}  & \blocket{{108}}{{108}}{{48}}{5} & \outsizesRaw{\tcolor{16}}{\xycolor{28}}{\xycolor{28}}{3} \\
				&  & \\
				&  & \\
				\hline
				\multirow{3}{*}{res$_4$}  & \blocket{{216}}{{216}}{{96}}{11} & \outsizesRaw{\tcolor{16}}{\xycolor{14}}{\xycolor{14}}{3} \\
				&  & \\
				&  & \\
				\hline
				\multirow{3}{*}{res$_5$}  & \blocket{{432}}{{432}}{{192}}{7} & \outsizesRaw{\tcolor{16}}{\xycolor{7}}{\xycolor{7}}{3}  \\
				&  & \\
				&  & \\
				\hline
				\multirow{1}{*}{conv$_5$} & \multirow{1}{*}{1\x1$^\text{2}$, {432}}   & \outsizesRaw{\tcolor{16}}{\xycolor{7}}{\xycolor{7}}{1}   \\
				\multirow{1}{*}{pool$_5$} &\outsizesRaw{{\tcolor{16}}}{\xycolor{7}}{\xycolor{7}}{1}   & 1\x1\x1    \\
				\multirow{1}{*}{fc$_1$} & \multirow{1}{*}{1\x1$^\text{2}$, {2048}}    & 1\x1\x1    \\
				\multicolumn{1}{c|}{ fc$_2$}  & \multirow{1}{*}{1\x1$^\text{2}$, { \#classes}} & 1\x1\x1  \\
		\end{tabular}}
	}\hspace{5pt}
	\subfloat[\textbf{X3D-XL} with 35.84G FLOPs \& 10.99M param, and 78.4\% top-1 acc.~using expansion of  \gatau $=$ 5, \gat $=$ 16, \mbox{\gaxy $=$ 2$\sqrt{\text{2}}$,} \gaw $=$ 2.9, \gab $=$ 2.25, \gad $=$ 5.
	\label{tab:arch_xl}]{
		\tablestyle{2pt}{1.05}
		\resizebox{0.31\linewidth}{!}{
			\tablestyle{1pt}{1.08}
			\begin{tabular}{c|c|c}
				stage & filters & output sizes $T$\x$H$\x$W$ \\
				\shline
				\multirow{1}{*}{data layer} & \multirow{1}{*}{stride 5, 1$^\text{2}$}  &  \outsizesRaw{16}{\xycolor{312}}{\xycolor{312}}{1}   \\
				\hline
				\multirow{1}{*}{conv$_1$} & \multirow{1}{*}{1\x3$^\text{2}$, 3\x1, \wcolor{32}}  & \outsizesRaw{16}{\xycolor{156}}{\xycolor{156}}{1}   \\
				\hline
				\multirow{3}{*}{res$_2$}  & \blocket{\wcolor{72}}{\wcolor{72}}{\wcolor{32}}{\dcolor{5}} & \outsizesRaw{16}{\xycolor{78}}{\xycolor{78}}{3}  \\
				&  & \\
				&  & \\
				\hline
				\multirow{3}{*}{res$_3$}  & \blocket{\wcolor{162}}{\wcolor{162}}{\wcolor{72}}{\dcolor{10}} & \outsizesRaw{16}{\xycolor{39}}{\xycolor{39}}{3} \\
				&  & \\
				&  & \\
				\hline
				\multirow{3}{*}{res$_4$}  & \blocket{\wcolor{306}}{\wcolor{306}}{\wcolor{136}}{\dcolor{25}} & \outsizesRaw{16}{\xycolor{20}}{\xycolor{20}}{3} \\
				&  & \\
				&  & \\
				\hline
				\multirow{3}{*}{res$_5$}  & \blocket{\wcolor{630}}{\wcolor{630}}{\wcolor{280}}{\dcolor{15}} & \outsizesRaw{16}{\xycolor{10}}{\xycolor{10}}{3}  \\
				&  & \\
				&  & \\
				\hline
				\multirow{1}{*}{conv$_5$} & \multirow{1}{*}{1\x1$^\text{2}$, \wcolor{630}}   & \outsizesRaw{16}{\xycolor{10}}{\xycolor{10}}{1}   \\
				\multirow{1}{*}{pool$_5$} &\outsizesRaw{{16}}{\xycolor{10}}{\xycolor{10}}{1}   & 1\x1\x1    \\
				\multirow{1}{*}{fc$_1$} & \multirow{1}{*}{1\x1$^\text{2}$, {2048}}    & 1\x1\x1    \\
				\multicolumn{1}{c|}{ fc$_2$}  & \multirow{1}{*}{1\x1$^\text{2}$, { \#classes}} & 1\x1\x1  \\
		\end{tabular}}
	}
	\vspace{-.8em}
	\caption{Three instantiations of {X3D} with varying complexity. The top-1 accuracy corresponds to \textit{10-Center} view testing on K400. The models in \protect\subref{tab:arch_s} and \protect\subref{tab:arch_m} only differ in spatiotemporal resolution of the input and activations (\gat, \gatau, \gaxy),  and  \protect\subref{tab:arch_xl}  differs from  \protect\subref{tab:arch_m} in spatial resolution, \gaxy, width, \gaw,  and depth, \gad. See Table~\ref{tab:arch} for X2D. Surprisingly X3D-XL has a maximum width of 630 feature channels. 
	}
	\label{tab:arch_instances}
	\vspace{-5pt}
\end{table*}

(i) First of all, expanding along \emph{any} one of the candidate axes increases accuracy. This justifies our motivation of taking multiple axes (instead of just the temporal axis) into account when designing spatiotemporal models.

(ii) Surprisingly, the first step selected by the expansion algorithm is \emph{not} along the temporal axis; instead, it is a factor that grows the ``bottleneck" width \gab~in the ResNet block design \cite{He2016}. This echoes the inverted bottleneck design in \cite{Sandler2018} (called ``inverted residual" \cite{Sandler2018}). This is possibly because these layers are lightweight (due to the channel-wise design of MobileNets) and thus are economical to expand at first. Another interesting observation is that accuracy varies strongly, with the bottleneck expansion \gab~providing the highest top-1 accuracy of 55.0\% and depth expansion \gad~the lowest with 51.3\% at same complexity of 41.4M FLOPs.

(iii) The second step extends the temporal size of the model from one to two frames (expanding \gatau~and \gat~is identical for this step as there exists only a single frame in the previous one). This is what we expected to be the most effective expansion already in the first step as it enables the network to model temporal information for recognition.

(iv) The third step increases the spatial resolution \gaxy~ and starts to show a pattern that is interesting. The expansion increases spatial and temporal resolution followed by depth (\gad) in the fourth step. This is followed by multiple temporal expansions that increase temporal resolution (\ie frame-rate) and input duration (\gatau~\&~\gat), followed by two more expansions across the spatial resolution, \gaxy,~in steps 8 and 9, while step 10 increases the depth of the network, \gad. An expansion of the depth after increasing input resolution is intuitive, as it allows to grow the filter receptive field resolution and size within each residual stage. 

(v) Even though we start from a base model that is intentionally made tiny by having very few channels, the expansion does \emph{not} choose to globally expand the width up to the 10$^\text{th}$ step of the expansion process, making X3D  similar to the Fast pathway design \cite{Feichtenhofer2019a} with high spatiotemporal resolution but low width. The last expansion step shown in the top-right of \figref{fig:expansion_curve} increases the width \gaw. The final two steps, not shown in \figref{fig:expansion_curve}, expand \gatau~and \gad.

In the spirit of VGG models \cite{Chatfield2014a,Simonyan2015} we define a set of networks based on their target complexity. We use FLOPs  as this reflects a hardware agnostic measure  of model complexity. Parameters are also possible, but as they would not be sensitive to the input and activation tensor size, we only report them as secondary metric. To cover the models from our expansion, \tblref{tab:expansion:x3d} defines complexity regimes by FLOPs, ranging from extra small (XS) to extra extra large (XXL).

\paragraph{Expanded instances.} The smallest instance, \textbf{X3D-XS} is the output after 5 expansion steps. Expansion is simple and efficient as it requires to train few models that are mostly at a low compute regime. For \textbf{X3D-XS} each step trains models of around 0.04, 0.08, 0.15, 0.30, 0.60 GFLOPs. Since we train one model for each of the 6 axes the approximate cost for these five steps is roughly equal to training a single model of 6 \x 1.17 GFLOPS (to be fair, this ignores overhead cost for data loading \etc as 6\x5$=$30 models are trained overall).

The next larger model is \textbf{X3D-S} which is defined by one backward \textit{contraction} step after the 7$^\text{th}$ expansion step. The contraction step simply reduces the expansion (\gat) proportionally to roughly match the target regime of $\leq$ 2 GFLOPs. For this model we also tried to contract each other axis to match the target and found that \gat~is best among the others. 

The next models in \tblref{tab:expansion:x3d} is \textbf{X3D-M} ($\leq$ 2 GFLOPs) that achieves 74.6\% top-1 accuracy, \textbf{X3D-L} ($\leq$ 20 GFLOPs) with 76.8\% top-1  and \textbf{X3D-XL} ($\leq$ 40 GFLOPs)  with 78.4\% and \textbf{X3D-XXL} ($\leq$ 150 GFLOPs)  with 80.0\%  top-1 accuracy by expansion in the consecutive steps. 

Further speed/accuracy comparisons are provided in \S\ref{sec:resultsapp}. 
\tblref{tab:arch_instances} shows three instantiations of X3D with varying complexity. It is interesting to inspect the differences of the models, X3D-S in \tblref{tab:arch_s} is just a lower spatiotemporal resolution (\gat, \gatau, \gaxy) version of \tblref{tab:arch_m}; therefore has the \textit{same number of parameters}, and  X3D-XL in \tblref{tab:arch_xl}  is created by expanding X3D-M \ref{tab:arch_m} in spatial resolution (\gaxy) and width (\gaw). See Table~\ref{tab:arch} for X2D.

	\subsection{Main Results}

\paragraph{Kinetics-400.} \tblref{tab:sota:k400} shows the comparison with state-of-the-art results for three X3D instantiations.To be comparable to previous work, we use the same testing strategy, that is \textit{10-LeftCenterRight} (\ie 30-view)
inference. For each model, the table reports (from-left-to-right) ImageNet pretraining (\textit{pre}), top-1 and top-5 validation accuracy, average test accuracy as \mbox{(top-1$+$ top-5)$/$2} (\ie official test-server metric), inference cost (GFLOPs\x views) and parameters. 

In comparison to the state-of-the-art, SlowFast \cite{Feichtenhofer2019}, \mbox{\textbf{X3D-XL}}, provides comparable (slightly lower) performance (-0.7\% top-1 and identical top-5 accuracy), while requiring 4.8\x~fewer multiply-add operations (FLOPs) and 5.5\x~fewer parameters than SlowFast 16\x 8, R101 + NL  blocks \cite{Wang2018}, and better accuracy than \mbox{SlowFast} 8\x 8, R101+NL with 2.4\x~fewer multiply-add operations and 5.5\x~fewer parameters.  When comparing \textbf{X3D-L}, we observe similar performance as Channel-Separated Networks (ip-CSN-152) \cite{Tran2019} and SlowFast 8\x 8, at 4.3\x~fewer FLOPs and 5.4\x~fewer parameters. Finally, in the lower compute regime \textbf{X3D-M} is comparable to SlowFast 4\x 16, R50 and Oct-I3D + NL \cite{chen2019drop} while having 4.7\x~fewer FLOPs and 9.1\x~fewer parameters. We observe consistent results on the \textit{test} set with the largest (and least efficient) \textbf{X3D-XXL} producing 86.7\% average top1$/$5 accuracy, showing good generalization performance.

\begin{table}[t!]
	\vspace{-14pt}
	\hspace*{-7pt}
	\centering
	\tablestyle{1.8pt}{1.05}
	\resizebox{1.04\linewidth}{!}{
		\begin{tabular}{l|c|c|c|c|c|r}
			\multicolumn{1}{c|}{model}  &\multicolumn{1}{c|}{pre} &   top-1  & top-5  & test & {\scriptsize GFLOPs\x views}  & Param \\
			\shline
			{I3D \cite{Carreira2017}} & \multirow{7}{*}{\rotatebox{90}{ImageNet}} & {71.1} &  {90.3} & 80.2 & {108 \x} {N/A} & 12M  \\
			{Two-Stream I3D \cite{Carreira2017}}  &  {} &   {75.7} &  {92.0} & 82.8 & {216~\x} {N/A} & 25M  \\
			{Two-Stream S3D-G \cite{Xie2018}}  &  {} &   {77.2} & {93.0} &&  {143~\x}  {N/A} & 23.1M  \\
			{MF-Net\cite{chen2018multi}}  &  {} &   {72.8} & {90.4} &  &{11.1~\x~50}  & 8.0M  \\
			TSM R50 \cite{lin2018temporal} &  & 74.7 &  N/A & &65 \x~10& 24.3M  \\
			{Nonlocal R50 \cite{Wang2018}}   &  {} &   {76.5} & {92.6} & & {282~\x}  {30} & 35.3M  \\
			{Nonlocal R101 \cite{Wang2018}}  &  {} &   {77.7} & {93.3} & 83.8 &{359~\x}  {30} & 54.3M \\
			
			\hline

			Two-Stream I3D \cite{Carreira2017}&   -&    71.6 & 90.0 & &216~\x~NA & 25.0M \\
			
			R(2+1)D \cite{Tran2018}&     -&   72.0 &  90.0 & &152~\x~115  & 63.6M \\
			Two-Stream R(2+1)D \cite{Tran2018}  &  -&  73.9 &  90.9& &304~\x~115 & 127.2M  \\
			Oct-I3D + NL \cite{chen2019drop} &    -&   75.7  & N/A && 28.9 \x~30 &  33.6M  \\
			
			ip-CSN-152  \cite{Tran2019}  &  -&  77.8 & 92.8 && 109~\x~30& 32.8M  \\
			
			{SlowFast} 4\x 16, R50 \cite{Feichtenhofer2019}    &- & 75.6  & 92.1 && 36.1~\x~30 & 34.4M \\

			{SlowFast} 8\x 8, R101 \cite{Feichtenhofer2019}  & - & 77.9  & 93.2 & 84.2 &  106~\x~30 & 53.7M  \\

				{SlowFast} 8\x 8, R101+NL \cite{Feichtenhofer2019}   &- & {78.7} & {93.5} & 84.9 &  116~\x~30 & 59.9M  \\ 

			{SlowFast} 16\x 8, R101+NL \cite{Feichtenhofer2019}    &- &{79.8} & {93.9} & {85.7} & 234~\x~30 & 59.9M  \\
			
			\hline
			\textbf{X3D-M}     &- & 76.0  & 92.3 & 82.9 &\textbf{ 6.2~\x~30}  & \textbf{3.8M}  \\
			\textbf{X3D-L}     &- & 77.5  & 92.9 & 83.8 & 24.8~\x~30  &	6.1M\\
			\textbf{X3D-XL}     &- & 79.1  & {93.9} & {85.3} &48.4~\x~30 & 11.0M  \\ %
				\textbf{X3D-XXL}     &- & \textbf{80.4}  & \textbf{94.6} & \textbf{86.7} & 194.1~\x~30 & 20.3M  \\ 
	\end{tabular}}
	\caption{\textbf{Comparison to the state-of-the-art on K400-val \& test}. We report the inference cost with a single ``view" (temporal clip with spatial crop) $\times$ the numbers of such views used (GFLOPs\x views). ``N/A'' indicates the numbers are not available for us. The ``test'' column  shows average of top1 and top5 on the Kinetics-400 testset.  
	}
	\label{tab:sota:k400}
	\vspace{-.2em}
\end{table}
\begin{table}[t]
	\centering
	\small
	\tablestyle{2pt}{1.05}
	\resizebox{1\linewidth}{!}{
		\begin{tabular}{l|c|c|c|c|r}
			\multicolumn{1}{c|}{model}  &\multicolumn{1}{c|}{pretrain} &   top-1  & top-5  & GFLOPs\x views  & Param \\
			\shline
			I3D \cite{Carreira2018} & - &   71.9 & 90.1  &  108~\x~N/A & 12M  \\
			Oct-I3D + NL  \cite{chen2019drop} & {\scriptsize ImageNet} &    {76.0} & {N/A} & {25.6~\x~30 } & 12M \\
			{SlowFast} 4\x 16, R50  \cite{Feichtenhofer2019} &  - & 78.8  & 94.0 & 36.1~\x~30 & 34.4M \\
			{SlowFast} 16\x8, R101+NL \cite{Feichtenhofer2019}  & - & {81.8}  & {95.1} &  234~\x~30 & 59.9M  \\
			\hline
			\textbf{X3D-M}     &- & 78.8  & 94.5 & \textbf{6.2~\x~30}  & \textbf{3.8M}  \\
			\textbf{X3D-XL} &   -  & \textbf{81.9}  & \textbf{95.5} & 48.4~\x~30 & 11.0M  \\
			
	\end{tabular}}
	\caption{\textbf{Comparison with the state-of-the-art on Kinetics-600}. Results are consistent with K400 in \tblref{tab:sota:k400} above.
	}
	\label{tab:sota:k600}
					\vspace{-10pt}
\end{table}

\paragraph{Kinetics-600} is a larger version of Kinetics that shall demonstrate further generalization of our approach. Results are shown in \tblref{tab:sota:k600}. Our variants demonstrate similar performance as above, with the best model now providing slightly better performance than the previous state-of-the-art SlowFast 16\x 8, R101+NL \cite{Feichtenhofer2019}, again for 4.8\x~fewer FLOPs (\ie multiply-add operations) and 5.5\x~fewer parameter. In the lower computation regime, \textbf{X3D-M} is comparable to  SlowFast 4\x 16, R50 but requires 5.8\x~fewer FLOPs and 9.1\x~fewer parameters. 

\paragraph{Charades} \cite{Sigurdsson2016} is a dataset with longer range activities. \tblref{tab:sota:charades} shows our results. 
\textbf{X3D-XL} provides higher performance (+0.9 mAP with K400 and +1.9mAP under K600 pretraining), while requiring 4.8\x~fewer multiply-add operations (FLOPs) and 5.5\x~fewer parameters than previous highest system, SlowFast \cite{Feichtenhofer2019} with+ NL blocks \cite{Wang2018}.

\subsection{Ablation Experiments}\label{sec:ablations}

	This section provides ablation studies on K400 val and test sets, comparing accuracy and computational complexity.

\paragraph{Comparison with EfficientNet3D.}  \label{sec:exp_efficient}

We first aim to compare X3D with a 3D extension of EfficientNet  \cite{tan2019efficientnet}. This architecture uses exactly the same implementation extras such as channel-wise separable convolution \cite{Howard2017} as  as X3D, but was found by searching a large number of models for optimal trade-off on image-classification.
This ablation studies if a direct extension of EfficientNet to 3D is comparable to X3D (which is expanded by only training few models).
 EfficientNet models are provided for various complexity ranges.  We ablate three versions, B0, B3 and B4 that are extended in 3D using uniform scaling coefficients \cite{tan2019efficientnet} for the spatial and temporal resolution.  

	\begin{table}[t!] 
		\vspace{-14.5pt}
	\centering
	\small
	\tablestyle{2pt}{1.05}
	\begin{tabular}{l|x{64}|x{22}|cr}
		\multicolumn{1}{c|}{model} &  \multicolumn{1}{c|}{pretrain} &  mAP    & \scriptsize GFLOPs\x views & Param \\   
		\shline
		Nonlocal \cite{Wang2018} &  {\scriptsize ImageNet+Kinetics400}  &   37.5 & 544~\x~30 & 54.3M  \\
		STRG, +NL \cite{Wang2018b} &  {\scriptsize ImageNet+Kinetics400} &   39.7 & 630~\x~30 & 58.3M \\
		Timeception \cite{hussein2019timeception} & {\scriptsize Kinetics-400} & 41.1 &  N/A\x N/A & N/A \\
		LFB, +NL \cite{Wu2019} & {\scriptsize Kinetics-400}  & 42.5 & 529 ~\x~30 & 122M    \\
		SlowFast, +NL \cite{Feichtenhofer2019}  & {\scriptsize Kinetics-400} & 42.5  &  234~\x~30  & 59.9M  \\
		SlowFast, +NL \cite{Feichtenhofer2019} &  {\scriptsize Kinetics-600} &  {45.2} &  234~\x~30  & 59.9M  \\ \hline
		\textbf{X3D-XL} &    {\scriptsize Kinetics-400} & \textbf{43.4} &  \textbf{48.4~\x~30} & \textbf{11.0M}  \\
		\textbf{X3D-XL} &    {\scriptsize Kinetics-600} & \textbf{47.1} &  \textbf{48.4~\x~30 }& \textbf{11.0M}  \\ 
	\end{tabular}
	\caption{\textbf{Comparison with the state-of-the-art on Charades}. SlowFast variants are based on  $T$\x$\tau$  $=$ 16\x8.} 	\vspace{-1em}
	\label{tab:sota:charades}
\end{table} 	

\newcommand{\pacc}[1]{{\bf \fontsize{7.5}{42}\selectfont \color{citecolor!80}~(#1)}}
\newcommand{\macc}[1]{{\bf \fontsize{7.5}{42}\selectfont \color{red!60}~(#1)}}
\begin{table}[t!]\centering  
		\vspace{10pt}
			\hspace*{-5pt}
	\resizebox{1.02\linewidth}{!}{
	\tablestyle{2.0pt}{1.1}
	\begin{tabular}{l|c|llll}
		\multicolumn{1}{c|}{\multirow{1}{*}{model}} &  \multirow{1}{*}{data} & \multirow{1}{*}{top-1} & \multirow{1}{*}{top-5} & FLOPs (G)  & Params (M) \\
		\shline
		EfficientNet3D-B0  & \multirow{5}{*}{K400}& 66.7   & 86.6 & 0.74 & 3.30  \\
		\textbf{X3D-XS}  &   \multirow{5}{*}{val}  & 68.6\pacc{$+$1.9} & 87.9\pacc{$+$1.3}  & 0.60\pacc{$-$1.4} & 3.76\macc{$+$0.5} \\    
		
		EfficientNet3D-B3  & \multirow{2}{*}{} & 72.4 & 89.6  &  6.91 &  8.19 \\ 
		\textbf{X3D-M}  &  &74.6\pacc{$+$2.2} & 91.7\pacc{$+$2.1}  &4.73\pacc{$-$2.2}  & 3.76\pacc{$-$4.4}  \\  
		
		EfficientNet3D-B4  & \multirow{2}{*}{} &  74.5 & 90.6  &  23.80 &  12.16 \\ 
		\textbf{X3D-L}  & &  76.8\pacc{$+$2.3}  & 92.5\pacc{$+$1.9}  & 18.37\pacc{$-$5.4}  & 6.08\pacc{$-$6.1}  \\  \hline
		
		EfficientNet3D-B0  & \multirow{5}{*}{K400}& 64.8   & 85.4 & 0.74 & 3.30  \\
		\textbf{X3D-XS}  &  \multirow{5}{*}{test} & 66.6\pacc{$+$1.8}  & 86.8\pacc{$+$1.4}  & 0.60\pacc{$-$1.4} & 3.76\macc{$+$0.5} \\    
		EfficientNet3D-B3  & \multirow{2}{*}{}  & 69.9 & 88.1  &  6.91 &  8.19 \\ 
		\textbf{X3D-M}  & & 73.0\pacc{$+$2.1}  & 90.8\pacc{$+$2.7}  &4.73\pacc{$-$2.2}  & 3.76\pacc{$-$4.4} \\   
		EfficientNet3D-B4  & \multirow{2}{*}{} & 71.8 & 88.9  &  23.80 &  12.16 \\ 
		\textbf{X3D-L}  & &  74.6\pacc{$+$2.8}  & 91.4\pacc{$+$2.5}  & 18.37\pacc{$-$5.4}  & 6.08\pacc{$-$6.1}  \\  \hline
	\end{tabular}}
	\caption{\textbf{Comparison to EfficientNet3D}: 
		We compare to a 3D version of EfficientNet on K400-val and test. 10-Center clip testing is used. \mbox{EfficientNet3D} has the same mobile components as X3D. 
	}
	\label{tab:ablations:efficientnet}
	\vspace{-0.8em}
\end{table}

In \tblref{tab:ablations:efficientnet}, we compare three X3D models of similar complexity to EfficientNet3D on two sets, K400-val and K400-test (from top-to-bottom). 
Starting with K400-val (top rows), our model \textbf{X3D-XS}, corresponding to only 4 expansion steps in \figref{fig:expansion_curve}. is comparable in FLOPs (slightly lower) and parameters (slightly higher), to  \mbox{EfficientNet3D-B0}, but achieves 1.9\% higher top-1 and 1.3\% higher top-1 accuracy.

Next, comparing  \textbf{X3D-M} to EfficientNet3D-B3 shows a gain of 2.0\% top-1 and 2.1\% top-5, despite using  32\% fewer FLOPs and 54\% fewer parameters. Finally, comparing \textbf{\mbox{X3D-L}} to EfficientNet3D-B4 shows a gain of 2.3\% top-1 and 1.9\% top-5, while having  23\% and 50\% fewer FLOPs and  parameters, respectively. Seeing larger gains for larger models underlines the benefit of progressive expansion, as more expansion steps have been performed for these.  

 Since our expansion is measured by validation set performance, it is interesting to see if this provides a benefit for X3D. Therefore, we investigate potential differences on the K400-test set, in the lower half of \tblref{tab:ablations:efficientnet}, where similar, even slightly {higher improvements} in accuracy can be observed when comparing the same models as above, showing that our models generalize well to the test set.

\begin{figure}[t]
	\vspace{-3.5em}
	\centering
				\includegraphics[width=0.9\linewidth]{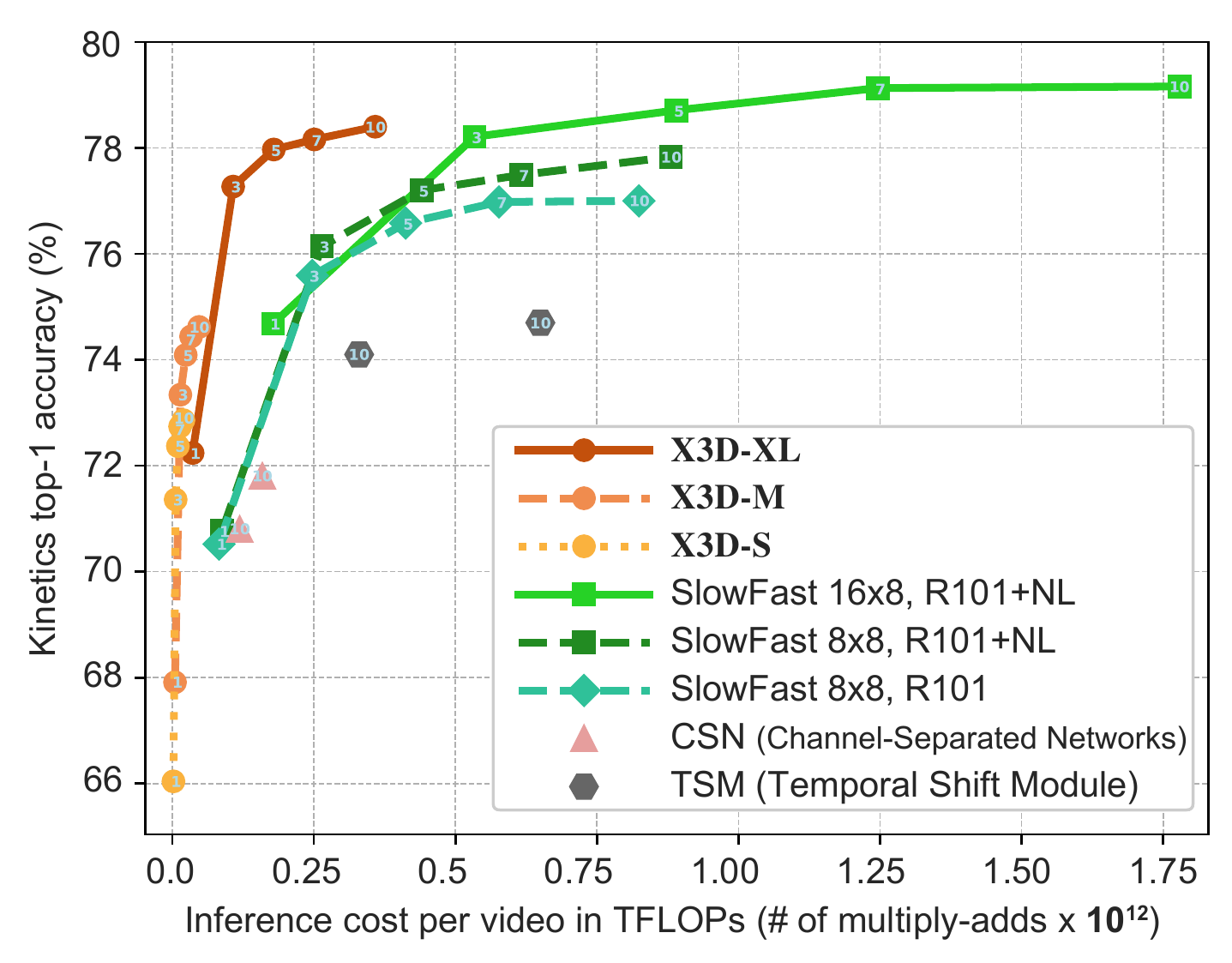}
	\vspace{-5pt}
	\caption{\textbf{Accuracy/complexity trade-off} on Kinetics-400 for different number of inference clips per video. The top-1 accuracy (vertical axis) is obtained by \clipscolor{$K$}-Center clip testing where the number of temporal clips $\clipscolor{K}\in \{\clipscolor{1,3,5,7,10}\}$ is shown in each curve. The horizontal axis shows the full inference cost per video. 
	}
	\label{fig:10clipTest}
	\vspace{-1.2em}
\end{figure}
\paragraph{Inference cost.} In many cases, like the experiments before, the inference procedure follows a \textit{fixed number of clips }for testing. Here, we aim to ablate the effect of using \textit{fewer }testing clips for video-level inference. In \figref{fig:10clipTest} we show the trade-off for the full inference of a video, when varying the number of temporal clips used. The vertical axis shows the top-1 accuracy on K400-{val}  and the horizontal axis the overall inference cost in FLOPs for different models.
Each model experiences a large performance increment when going from $K=$ 1 clip to 3-clip testing (which triples the FLOPs); this is expected as the 1-clip only covers the temporal center of an input video, while 3-clip covers start, center and end. Increasing the number of clips beyond 3 only marginally increases performance, signaling that efficient video inference can be performed with \textit{sparse clip sampling} if highest accuracy is not crucial. Lastly, when comparing different models we observe that X3D architectures can achieve similar accuracy as SlowFast \cite{Feichtenhofer2019}, CSN \cite{Tran2019} or TSM \cite{lin2018temporal} (for the latter two, only 10-clip testing results are available to us), while requiring 3-20\x~fewer multiply-add operations. Notably, the SlowFast 16\x8 variant does not benefit from increasing 7 to 10 temporal clips, showing that the expensive per clip cost of longer-term models does not reflect full inference efficiency as they allow sparser temporal sampling.  
A log-scale version of \figref{fig:10clipTest} and similar plots on K400-{test} are in \S\ref{sec:resultsapp}.

	\section{Experiments: AVA Action Detection}\label{sec:detection}

\paragraph{Dataset.}
The AVA dataset \cite{Gu2018} comes with bounding box annotations for spatiotemporal localization of (possibly multiple) human actions. There are 211k training and 57k validation video segments. We follow the standard protocol reporting mean Average Precision (mAP) on 60 classes \cite{Gu2018}.
\noindent\textbf{Detection architecture.}
We exactly follow the detection architecture in \cite{Feichtenhofer2019} to allow direct comparison of X3D with SlowFast networks as a backbone.
The detector is similar to Faster R-CNN \cite{Ren2015} with minimal modifications adapted for video. Details on implementation and training are in \S\ref{sec:detectionapp}.

\paragraph{Inference.} We perform inference on a single clip with \gat~frames sampled with stride \gatau~centered at the frame that is to be evaluated. Spatially we use a single center crop of 128\gaxy\x128\gaxy~pixels as in \cite{Wu2019}, to have a comparable measure for overall test costs, since fully-convolutional inference   has variable cost depending on the input video size.  
 
	\subsection{Main Results}

	We compare with state-of-the-art methods on AVA in \tblref{tab:ava:sota}.  To be comparable to previous work, we report results on the older AVA version 2.1 and newer 2.2 (which provides more consistent annotations), for our models pre-trained on K400 or K600. We compare against long-term feature banks (LFB) \cite{Wu2019} and SlowFast \cite{Feichtenhofer2019} as these are state-of-the art and use the same detection architecture as ours, varying the backbone from LFB, SlowFast and X3D. Note there are other recent works on AVA \eg, \cite{Sun2018,Girdhar2019,zhang2019structured,yang2019step}.
	
	 In the upper part of \tblref{tab:ava:sota} we compare \textbf{X3D-XL} with LFB, that uses a heavy backbone architecture for short and long-term modeling.  X3D-XL provides comparable accuracy (+0.3 mAP \vs LFB R50 and -0.7mAP \vs LFB R101) at greatly reduced cost by 10.9\x$/$14\x~fewer multiply-adds and 6.7\x$/$11.1\x~fewer parameters than  LFB R50$/$R101.
	
	Comparing to SlowFast \cite{Feichtenhofer2019} in the lower portion of the table we observe that \textbf{X3D-M} is lower than {SlowFast} 4\x16, R50 by 1.5mAP, but requiring 8.5\x~less multiply-adds and 10.9\x less parameters for this result. Comparing the larger \textbf{X3D-XL} to SlowFast 8\x8 + NL we observe the same performance at 3\x~and 5.4\x~fewer multiply-adds and parameters.

\begin{table}[t!]
	\vspace{-2.2em} 
	\hspace*{-7pt}
	\centering
	\tablestyle{2.5pt}{1.05}
	\begin{tabular}{l|c|c|c|c|r}
		\multicolumn{1}{c|}{model} & AVA &\multicolumn{1}{c|}{pre} &   val mAP  & GFLOPs & Param \\ 
		\shline
		
		LFB, R50+NL  \cite{Wu2019} &  \multirow{3}{*}{v2.1}  & \multirow{3}{*}{K400}  & 25.8 & 529 & 73.6M   \\
		
		LFB, R101+NL \cite{Wu2019} & & & 26.8 & 677 & 122M \\

		\textbf{X3D-XL}   &   &  &  26.1 & 48.4 & 11.0M \\

		\hline
		
		{SlowFast} 4\x16, R50 \cite{Feichtenhofer2019} &  \multirow{4}{*}{v2.2} &  \multirow{4}{*}{K600}   & 24.7 & 52.5 & 33.7M \\
		{SlowFast}, 8\x8 R101+NL \cite{Feichtenhofer2019}  &  &   &  27.4 & 146 & 59.2M \\ 

		\textbf{X3D-M} & &    & 23.2 & \textbf{6.2 }& \textbf{3.1M} \\
		\textbf{X3D-XL} & &    &  \textbf{27.4} & 48.4 & 11.0M \\

	\end{tabular}
	\caption{\textbf{Comparison with the state-of-the-art on AVA}. All methods use \textit{single center crop} inference; full testing cost is directly proportional to the the floating point operations (GFLOPs) by multiplying with  the number of validation segments (57k) in the dataset. }
	\label{tab:ava:sota}
	\vspace{-1em}
\end{table}

\section{Conclusion}

This paper presents X3D, a spatiotemporal architecture that is progressively expanded from a tiny spatial network. Multiple candidate axes, in space, time, width and depth are considered for expansion under good computation/accuracy trade-off. A surprising finding of our progressive expansion is that networks with thin channel dimension and high spatiotemporal resolution can be effective for video recognition. 
X3D achieves competitive efficiency, and we hope that it can foster future research and applications in video recognition.

		\paragraph{Acknowledgements:} I thank Kaiming He, Jitendra \mbox{Malik},  Ross Girshick, and Piotr Doll\'ar for valuable discussions and \mbox{encouragement}.

	\cleardoublepage

\newcount\cvprrulercount
\appendix
\section*{Appendix}

\setcounter{table}{0}
\renewcommand{\thetable}{A.\arabic{table}}
\renewcommand{\thefigure}{A.\arabic{figure}}

This appendix provides further details as referenced in the main paper: 
\sref{sec:training} contains additional implementation details  for: AVA Action Detection  (\S\ref{sec:detectionapp}),  Charades Action Classification (\S\ref{sec:charades}), and  Kinetics Action Classification (\S\ref{sec:kineticsapp}).  \sref{sec:resultsapp} contains further results and ablations on Kinetics-400.

\section{Additional Implementation Details} \label{sec:training} 

\subsection{Details: AVA Action Detection}\label{sec:detectionapp}

\paragraph{Detection architecture.}
We exactly follow the detection architecture in \cite{Feichtenhofer2019} to allow direct comparison of X3D with SlowFast networks as a backbone.
The detector is similar to Faster R-CNN \cite{Ren2015} with minimal modifications adapted for video. 
Since our paper focuses on efficiency, by default, we do not increase the spatial resolution of res$_5$ by 2$\times$ \cite{Feichtenhofer2019}. 
Region-of-interest (RoI) features \cite{Girshick2015} are extracted at the last feature map of res$_5$ by extending a 2D proposal at a frame into a 3D RoI by replicating it along the temporal axis, similar as done in previous work \cite{Gu2018,Sun2018,Jiang2018}, followed by application of frame-wise RoIAlign \cite{He2017}  and temporal global average pooling. The RoI features are then max-pooled and fed to a per-class, sigmoid classifier for prediction.

\paragraph{Training.} For direct comparison, the training procedure and hyper-parameters for AVA follow \cite{Feichtenhofer2019} without modification. 
The network weights are initialized from the Kinetics models and we use step-wise learning rate decay, that is reduced by 10\x~when validation error saturates. 
We train for 14k iterations (68 epochs for $\app$211k data), with linear warm-up \cite{Goyal2017} for the first 1k iterations and use a weight decay of 10$^{-7}$, as in \cite{Feichtenhofer2019}. 
All other hyper-parameters are the same as in the Kinetics experiments.
Ground-truth boxes, and proposals overlapping with ground-truth boxes by \mbox{IoU $>$ 0.9}, are used as the samples for training. The inputs are instantiation-specific clips of size \gat\x112\gaxy\x112\gaxy with time stride \gatau.

The region proposal extraction also follows \cite{Feichtenhofer2019} and is summarized here for completeness. 
We follow previous works that use pre-computed proposals \cite{Gu2018,Sun2018,Jiang2018}. Our region proposals are computed by an off-the-shelf person detector, \ie, that is not jointly trained with the action detection models.
We adopt a person-detection model trained with \emph{Detectron} \cite{Detectron2018}. It is a Faster R-CNN with a ResNeXt-101-FPN~\cite{Xie2017,Lin2017} backbone.
It is pre-trained on ImageNet and the COCO human keypoint images~\cite{Lin2014}.
We fine-tune this detector on AVA for person (actor) detection.
The person detector produces 93.9 AP@50 on the AVA validation set.
Then, the region proposals for action detection are detected person boxes with a confidence of $>$ 0.8, which has a recall of 91.1\% and a precision of 90.7\% for the person class.

\subsection{Details: Charades Action Classification}\label{sec:charades}

For \textbf{Charades}, we fine-tune the Kinetics models. All settings are the same as those of Kinetics, except the following. A per-class sigmoid output is used to account for the mutli-class nature. We train on a single machine for 24k iterations using a batch size of 16 and a base learning rate of 0.02 with 10\x~step-wise decay if the validation error saturates. We use weight decay of 10$^\text{-5}$. We also increase the model temporal stride by \x2~ as this dataset benefits from longer clips. For inference, we temporally max-pool prediction scores \cite{Wang2018, Feichtenhofer2019}.

\subsection{Details: Kinetics Action Classification}\label{sec:kineticsapp}

\paragraph{Training details.}
We use the initialization in \cite{He2015}. We adopt synchronized SGD training on 128 GPUs following the recipe in \cite{Goyal2017}.
The mini-batch size is 8 clips per GPU (so the total mini-batch size is 1024). 
We train with Batch Normalization (BN) \cite{Ioffe2015}, and the BN statistics are computed within each 8 clips, unless noted otherwise. 
We adopt a half-period cosine schedule \cite{Loshchilov2016} of learning rate decaying: the learning rate at the $n$-th iteration is $\eta\cdot0.5[\cos(\frac{n}{n_\text{max}}\pi)+1]$, where $n_\text{max}$ is the maximum training iterations and the base learning rate $\eta$ is set as 1.6.
We also use a linear warm-up strategy \cite{Goyal2017} in the first 8k iterations.
Unless specified, we train for 256 epochs (60k iterations with a total mini-batch size of 1024, in $\app$240k Kinetics videos).
We use momentum of 0.9, weight decay of 5\x10$^\text{-5}$ and dropout \cite{Hinton2012b} of 0.5 is used before the final classifier. 

For \textbf{Kinetics-600}, we extend the training epochs (and schedule) of above by 2\x. All other hyper-parameters are exactly as for Kinetics-400.

\paragraph{Implementation details.}

Non-Local (NL) blocks \cite{Wang2018} are not used for X3D. For SlowFast results, we use exactly the same implementation details as in \cite{Feichtenhofer2019}. Specifically, for SlowFast models involving NL, we initialize them with the counterparts that are trained without NL, to facilitate convergence. We only use NL on the (fused) Slow features of res$_4$ (instead of res$_3$+res$_4$ \cite{Wang2018}). 
For X3D and EfficientNet3D, we follow previous work on 2D mobile architectures \cite{tan2019mnasnet, tan2019efficientnet, Howard2019}, using SE blocks \cite{hu2018squeeze} (also found beneficial for efficient video classification in \cite{Xie2018}) and swish non-linearity \cite{ramachandran2017searching}. To conserve memory, we use SE with original reduction ratio of 1$/$16 only in every other residual block after the 3\x 3$^\text{2}$ conv; swish is only used before and after these layers and all other weight layers are followed by ReLU non-linearity  \cite{Krizhevsky2012}. We do not employ the ``linear-bottleneck'' design used in mobile image networks  \cite{Sandler2018, tan2019mnasnet, tan2019efficientnet, Howard2019}, as we found it to sometimes cause instability in distributed training, as it does not allow to zero-initialize the final BN scaling \cite{Goyal2017} of residual blocks.

\paragraph{Expansion details.}
To expand the model specified in \tblref{tab:arch}, we set all initial expansion factors, $\mathcal{X}_0$, to one \ie  \gat$=$\gaxy$=$\gaw$=$\gab$=$\gad$=$1 resulting in the X2D base model. A temporal sampling rate \gatau~is not defined for the X2D model as it does not have multiple frames. The smallest possible common expansion for this model is defined by increasing the number of frames from 1 to two; therefore we set the expansion-rate $\hat{c}$ to match the cost of increasing the temporal input length of the model by a factor of two (the smallest possible common increase in the first expansion step), which roughly doubles the cost of the model, $\hat{c}=2$.

Then, in every step of our expansion we train $a$ models, one for expanding each axis, such that its complexity doubles ($\hat{c}=2$). For the individual axes this roughly\footnote{The exact expansion factors slightly vary across steps to match the complexity increase, $\hat{c}$ (which is observable in Fig.~{\color{red} 2} of the main paper).} equals to the following operations:  

\begin{itemize}
	\item	\fastcolor{X-Fast:} \gatau $\leftarrow$ 0.5\gatau, reduces the sampling stride to double frame-rate while sampling the same input duration, this doubles the temporal size \gat$\leftarrow$2\gat.
	
	\item	\tcolor{X-Temporal:} Increases frame-rate by \mbox{\gatau $\leftarrow$ 0.75\gatau}~\textit{and} input duration to double the input size \gat$\leftarrow$2\gat~ (\ie 1.5\x~ higher frame-rate and 1.5\x longer input duration).
	
	\item	\xycolor{X-Spatial:} Expands the spatial resolution proportionally \gaxy $\leftarrow \sqrt{2}$\gaxy.
	
	\item	 \dcolor{X-Depth:} Expands the depth of the network by around  \gad $\leftarrow 2.2$\gad.
	
	\item	  \wcolor{X-Width:} Expands the global width for all layers by \gaw  $\leftarrow 2$\gaw. 
	
	\item	  \eicolor{X-Bottleneck:} Expands the bottleneck width by roughly \gab  $\leftarrow 2.25$\gab. 
\end{itemize}

The exact scaling factors slightly differ from one expansion step to the other due to rounding effects in network geometry (layers stride, activation size \etc). 

Since the stepwise expansion also allows to elegantly integrate regularization (which is typically increased for larger models), we perform a regularization expansion if the training error of the previous expansion step starts to deviate from the validation error. Specifically, we start the expansion with double the batch-size and half learning schedule than described above, then the BN statistics are computed within each 16 clips which lowers regularization and improves performance on small models. The batch-size is then decreased by 2\x~at the 8$^\text{th}$ step of expansion which increases generalization. We perform another regularization expansion at the 11$^\text{th}$/13$^\text{th}$ step by using drop-connect with rate 0.1/0.2 \cite{huang2016deep}.

\begin{figure*}[t]
	\vspace{-.5em}
	\centering

	\includegraphics[width=0.49\linewidth]{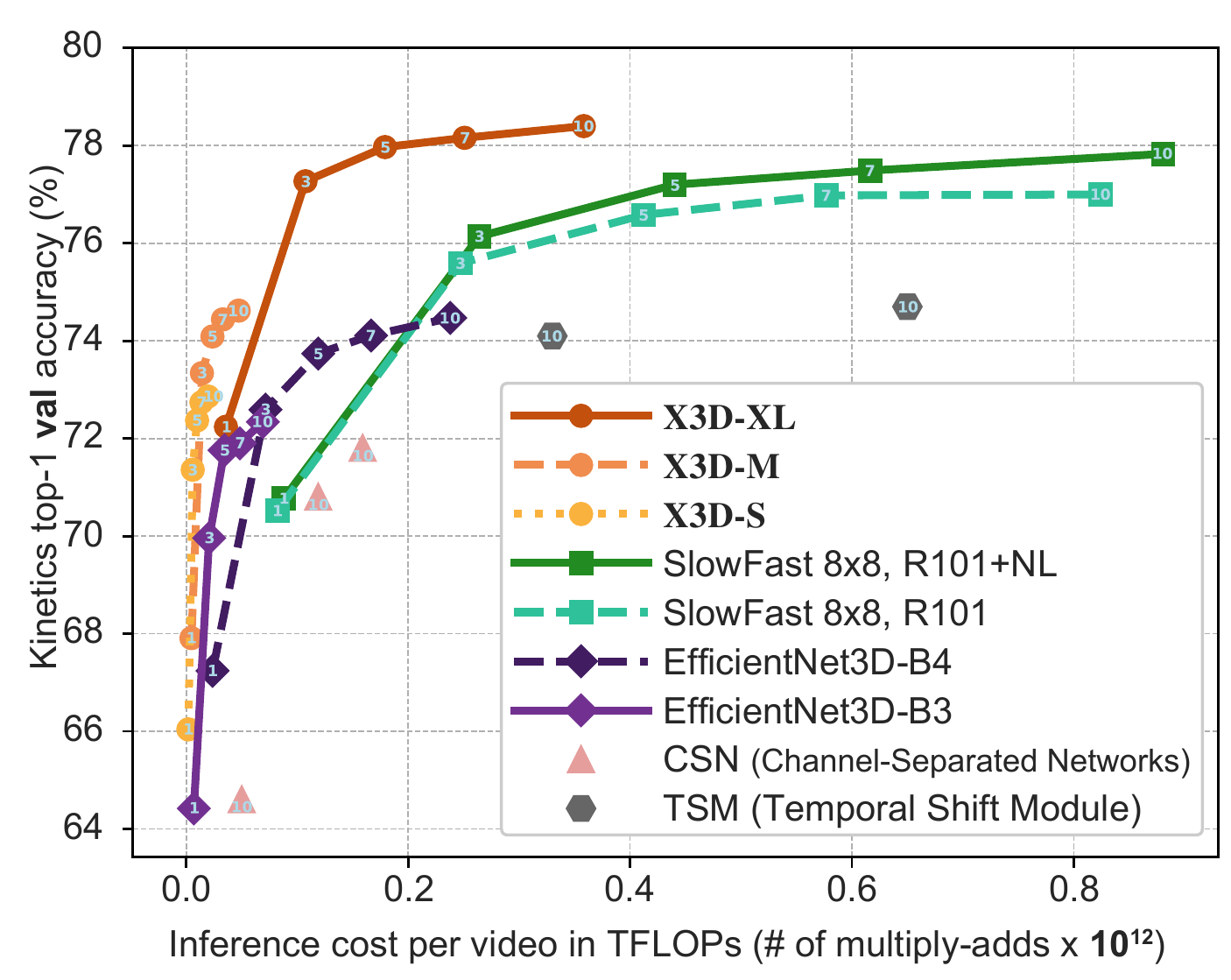}
	\includegraphics[width=0.49\linewidth]{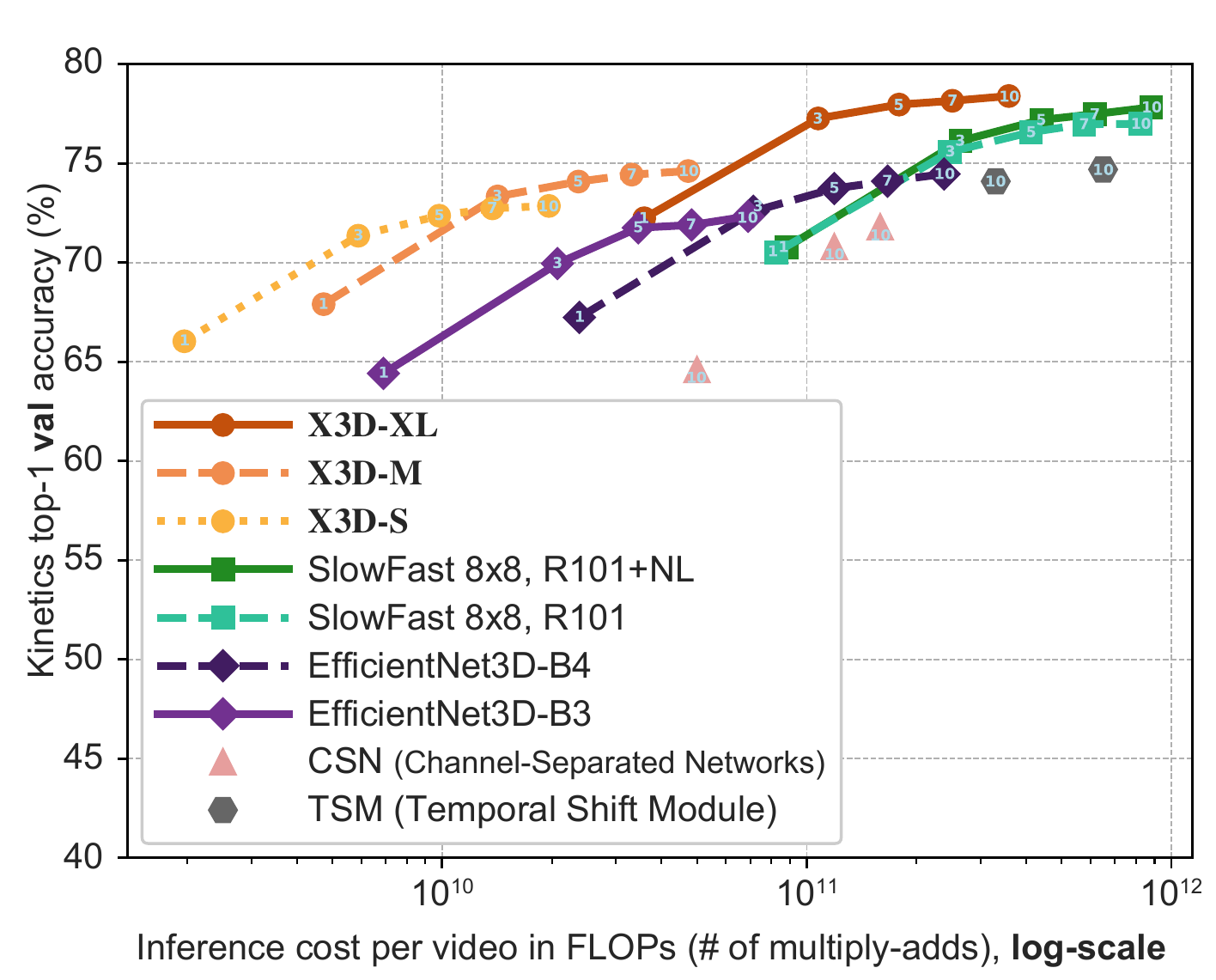}
	\includegraphics[width=.49\linewidth]{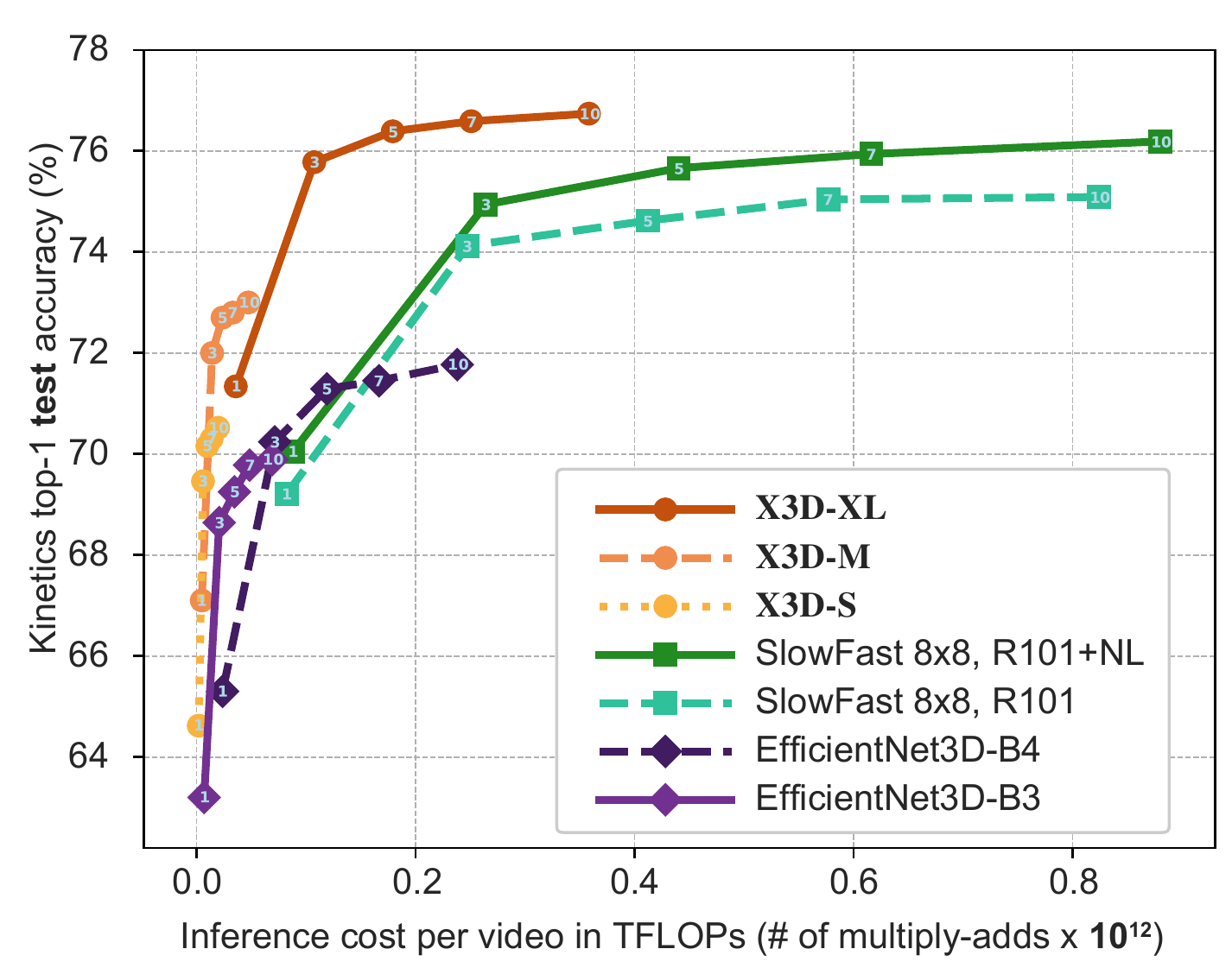}
	\includegraphics[width=0.49\linewidth]{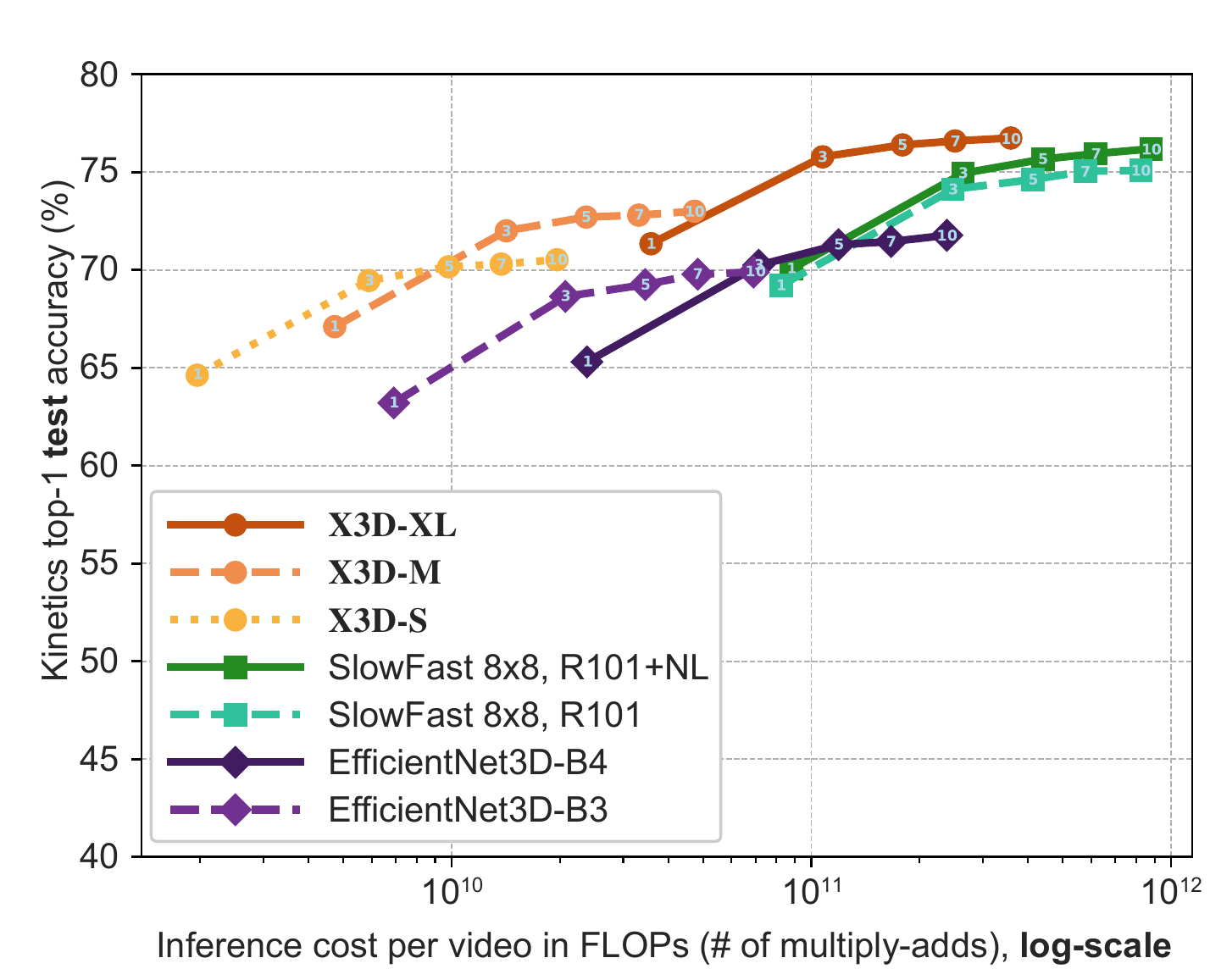}
			\vspace{5pt}
	\caption{\textbf{Accuracy/complexity trade-off} on K400-\textbf{val} (top) \& \textbf{test} (bottom) for varying  \# of inference clips per video. The top-1 accuracy (vertical axis) is obtained by \clipscolor{$K$}-Center clip testing where the number of temporal clips $\clipscolor{K}\in \{\clipscolor{1,3,5,7,10}\}$ is shown in each curve. The horizontal axis measures the full inference cost per video. The left-sided plots show a linear and the right plots a logarithmic (\textbf{log}) scale.
	}
	\label{fig:10clipTestAll}
\end{figure*}

\section{Additional Results} \label{sec:resultsapp}

\figref{fig:10clipTestAll} shows a series of extra plots on Kinetics-400, analyzed next (this extends Sec.~{\color{red} 4} of the main paper):

\paragraph{Inference cost.} Here we aim to provide further ablations for the effect of using \textit{fewer }testing clips for efficient video-level inference. In \figref{fig:10clipTestAll} we show the trade-off for the full inference of a video, when varying the number of temporal clips used. The vertical axis shows the top-1 accuracy on K400-{val}  and the horizontal axis the overall inference cost in FLOPs for different models.  

First, for comparison, the plot on top-left is the same as the one shown in Fig.~{\color{red} 3}. The plot on top-right shows this same plot with a logarithmic scale applied to the FLOPs axis. Using this scaling it is clearer to observe that smaller models ({X3D-S} and {X3D-M}) can provide up to 20\x~reduction in terms of multiply-add operations used during inference. 

For example, 3-clip {X3D-S} produces 71.4\% \mbox{top-1} at 5.9 GFLOPs, whereas 10-clip CSN-50 \cite{Tran2019} produces 70.8 \mbox{top-1} at 119 GFLOPs (20.2\x~higher cost), or 10-clip {X3D-S} 72.9\% \mbox{top-1} at 19.6 GFLOPs,  and 10-clip CSN-101 \cite{Tran2019} 71.8\% \mbox{top-1} at 159 GFLOPs (8.1\x~higher cost).

The lower two plots in \figref{fig:10clipTestAll} show the identical results on the test set of Kinetics-400 (which has been publicly released with Kinetics-600 \cite{Carreira2018}). Note that the test set is more challenging which leads to overall lower accuracy for all approaches \cite{ActivityNet2017}. We observe consistent results on the test set, illustrating good generalization of the models.

\textbf{Mobile components.}
Finally, we ablate the effect of mobile components employed in X3D and EfficientNet3D. Since the components can have different effects of models from the small and large computation regime, we ablate the effects on a small (X3D-S) and a large model (X3D-XL).  

First, we ablate channel-wise separable convolution \cite{Howard2017}, a key component in mobile ConvNets. We ablate two versions: (i) A version that reduces the bottleneck ratio (\gab) accordingly, such that the overall architecture preserves the multiple-add operations (FLOPs), and (ii) a version that keeps the originally, expanded bottleneck ratio. 

\tblref{tab:ablations_mobile} shows the results.  
For case (i) we see that performance drops significantly by 4\% top-1 accuracy for X3D-S and by 2.4\% for X3D-XL. For case (ii), we see that the performance of the baselines increases by 0.3\% and 0.8\% top-1 accuracy for X3D-S and X3D-XL, respectively. This shows that separable convolution is important for small-computation budgets, however, for best-performance a non-separable convolution can provide gains (at high cost).

Second, we ablate swish non-linearities \cite{ramachandran2017searching} (that are only implemented before and after the ``bottleneck'' convolution, to conserve memory). We observe that removing swish has a smaller performance decrease of 0.9\% for X3D-S and 0.4\% for X3D-XL, and therefore could  be changed to ReLU (which can be implemented in-place) if memory is priority. 

Third, we ablate SE blocks \cite{hu2018squeeze} (that are only used in every other residual block, to conserve memory). We observe that removing SE has a larger effect on performance, decreasing accuracy by 1.6\% for  X3D-S and 1.3\% for \mbox{X3D-XL}. \\These observed effects on performance are similar to the ones that have been shown Non-local (NL) attention blocks \cite{Wang2018}, and also in line with \cite{Xie2018}, where SE attention blocks have been found beneficial for efficient video classification.

\begin{table*}[t]\centering\vspace{-1em}
	\captionsetup[subfloat]{captionskip=2pt}
	\captionsetup[subffloat]{justification=centering}
	\subfloat[Ablating {mobile components on a Small model.}
	\label{tab:ablation:extras_s}]{
		\tablestyle{2pt}{1.05}
		\begin{tabular}{r|x{22}x{22}x{36}x{36}}
			model	 & top-1 & top-5 & FLOPs (G) & Param (M) \\
			\shline
			\multicolumn{1}{l|}{X3D-S}& 72.9 & 90.5 & 1.96 & 3.76 \\ 
			\hline
			$-$ CW conv \gab $=$ 0.6   & {68.9} & {88.8} & 1.95 & 3.16 \\ 
			$-$ CW conv & 73.2 & 90.4 & 17.6 & 22.1 \\ 
			$-$ swish & 72.0 & 90.4 & 1.96 & 3.76 \\ 
			$-$ SE  & 71.3 & 89.9 & 1.96 & 3.60 \\ 
	\end{tabular}}\hspace{3mm}
	\subfloat[Ablating {mobile components on an X-Large model.}
	\label{tab:ablation:extras_xl}]{
		\tablestyle{2pt}{1.05}
		\begin{tabular}{r|x{22}x{22}x{36}x{36}}
			model	 & top-1 & top-5 & FLOPs (G) & Param (M) \\
			\shline
			\multicolumn{1}{l|}{X3D-XL} & 78.4 & 93.6& 35.84 & 11.0\\ 
			\hline
			$-$ CW conv, \gab $=$ 0.56   & {76.0} & 92.6 & 34.80 & 9.73 \\
			$-$ CW conv & 79.2 & 93.5 & 365.4 & 95.1 \\
			$-$ swish & 78.0 & 93.4 & 35.84 & 11.0 \\ 
			$-$ SE  & 77.1 & 93.0 & 35.84 & 10.4 \\ 
	\end{tabular}}\hspace{3mm}
	
	\vspace{1em}
	\caption{{Ablations} of mobile components for video classification on {K400-val}. We show top-1 and top-5 classification accuracy (\%), parameters, and computational complexity measured in GFLOPs (floating-point operations, in \# of multiply-adds \x $10^9$) for a  single clip input of size \gat\x112\gaxy\x112\gax. Inference-time computational cost is reported GFLOPs \x10, as a fixed number of  \textit{10-Center} views is used. The results show that removing channel-wise separable convolution (CW conv) with unchanged bottleneck expansion ratio, \gab, drastically increases mutliply-adds and parameters at slightly higher accuracy, while swish has a smaller effect on performance than SE.  }
	\label{tab:ablations_mobile}
\end{table*}

{
	\small
	\bibliographystyle{ieee_fullname}
	\bibliography{x3d}
}
\end{document}